%% file: main.tex
\documentclass[letterpaper]{article} 
\usepackage{aaai2026}  
\usepackage{times}  
\usepackage{helvet}  
\usepackage{courier}  
\usepackage[hyphens]{url}  
\usepackage{graphicx} 
\usepackage{natbib}  
\usepackage{caption} 
\definecolor{RefBlue}{RGB}{20,38,126}
\usepackage[colorlinks=true,linkcolor=RefBlue,citecolor=RefBlue,urlcolor=RefBlue]{hyperref}
\usepackage{booktabs}       
\usepackage{amsfonts}       
\usepackage{nicefrac}       
\usepackage{microtype}      
\usepackage{xcolor}         
\usepackage{graphicx}
\usepackage{amsmath}
\usepackage{overpic}
\usepackage{color}
\usepackage{xcolor}
\usepackage{colortbl}
\usepackage{enumitem}
\usepackage{amsmath}
\usepackage{array}
\usepackage{multirow}
\usepackage{makecell}
\usepackage{amsmath} 
\usepackage{algorithm}
\usepackage{algorithmic}
\usepackage{capt-of} 
\usepackage{wrapfig}
\usepackage{newfloat}
\usepackage{listings}
\DeclareCaptionStyle{ruled}{labelfont=normalfont,labelsep=colon,strut=off} 
\floatstyle{ruled}
\newfloat{listing}{tb}{lst}{}
\floatname{listing}{Listing}
\newcommand{\etal}{\textit{et al.}}
\newcommand{\ie}{\textit{i.e.}}
\newcommand{\eg}{\textit{e.g.}}
\nocopyright
\title{Arbitrary-Scale 3D Gaussian Super-Resolution}  

\author {
    Huimin Zeng\textsuperscript{\rm 1}\thanks{Corresponding author.}, Yue Bai\textsuperscript{\rm 1}, Yun Fu\textsuperscript{\rm 1,\rm 2}
}
\affiliations {
    \textsuperscript{\rm 1}Department of Electrical and Computer Engineering, Northeastern University\\
    \textsuperscript{\rm 2}Khoury College of Computer Science, Northeastern University\\
    \{zeng.huim, bai.yue, y.fu\}@northeastern.edu
}

\usepackage{bibentry}

\usepackage{capt-of} 

\makeatletter
\renewcommand{\maketitle}{%
  \par%
  \begingroup 
    \def\thefootnote{\fnsymbol{footnote}}%
    \twocolumn[{%
      \@maketitle
      \vspace{-0.2in}%
      \begin{center}
        \includegraphics[
          width=1\textwidth,
          clip=true,
          trim=2.2cm 0.5cm 0 0.2cm
        ]{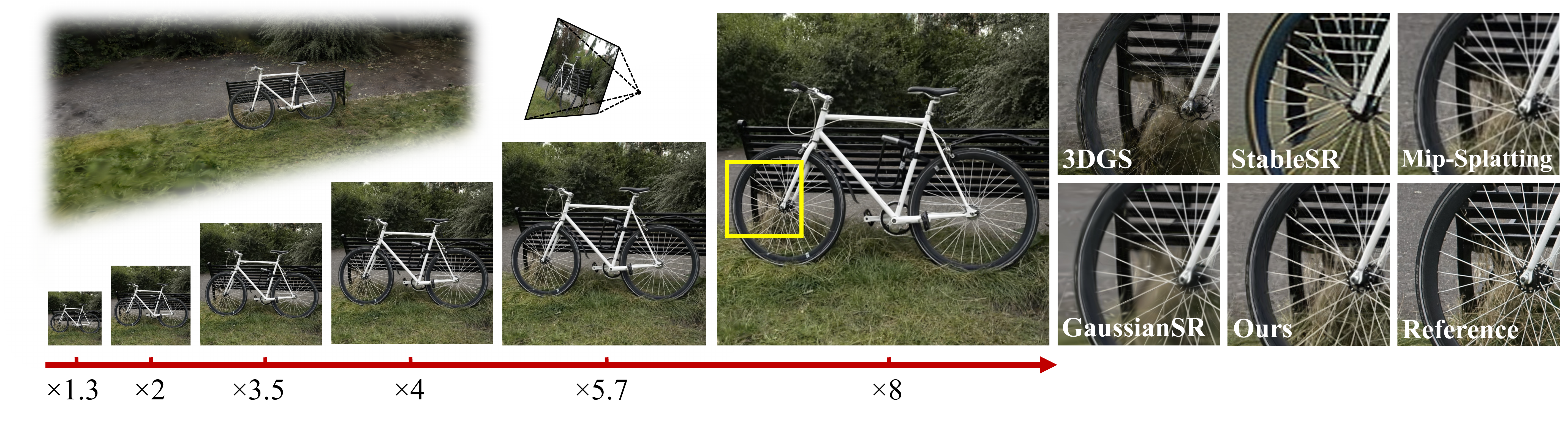}%
        \vspace{-8pt}%
        \captionof{figure}{%
          Visual results of typical solutions for arbitrary-scale 3D Gaussian super-resolution.
          Continuously rendering high-resolution novel views of different scale factors with vanilla 3DGS
          leads to aliasing artifacts. Cascaded solutions produce altered contents (\eg, StableSR).
          Anti-aliasing Mip-Splatting and GaussianSR yield limited details.%
        }%
        \label{fig:teaser}%
      \end{center}
      \vspace{0.1in}%
    }]%
    \@thanks%
  \endgroup%
  \if T\copyright@on
    \insert\footins{\noindent\footnotesize\copyright@text}%
  \fi%
  \setcounter{footnote}{0}%
  \let\maketitle\relax%
  \let\@maketitle\relax%
  \gdef\@thanks{}%
  \gdef\@author{}%
  \gdef\@title{}%
  \let\thanks\relax%
}%
\makeatother

\begin{document}

\maketitle

\begin{abstract}
Existing 3D Gaussian Splatting (3DGS) super-resolution methods typically perform high-resolution (HR) rendering of fixed scale factors, making them impractical for resource-limited scenarios. Directly rendering arbitrary-scale HR views with vanilla 3DGS introduces aliasing artifacts due to the lack of scale-aware rendering ability, while adding a post-processing upsampler for 3DGS complicates the framework and reduces rendering efficiency. To tackle these issues, we build an integrated framework that incorporates scale-aware rendering, generative prior-guided optimization, and progressive super-resolving to enable 3D Gaussian super-resolution of arbitrary scale factors with a single 3D model. Notably, our approach supports both integer and non-integer scale rendering to provide more flexibility. Extensive experiments demonstrate the effectiveness of our model in rendering high-quality arbitrary-scale HR views (\textbf{6.59 dB} PSNR gain over 3DGS) with a single model. It preserves structural consistency with LR views and across different scales, while maintaining real-time rendering speed (\textbf{85 FPS at 1080p}).
\end{abstract}

\begin{links}
    \link{Code}{https://github.com/huimin-zeng/Arbi-3DGSR}
\end{links}

\section{Introduction}\label{sec:intro}
High-resolution novel view synthesis (HRNVS) focuses on reconstructing 3D models from low-resolution (LR) sparse views and rendering high-resolution (HR) novel views.
Such capability is critical for resource-limited scenarios such as low-bandwidth streaming~\cite{grant2023meta,solmaz2020integration} and AR/VR applications on portable devices~\cite{fu2023auto}, where only images with limited size can be transferred and efficiently processed.
Previous Neural Radiance Field (NeRF) based methods~\cite{cross,supernerf,Huang_2023_CVPR,nerfsr,huang2024assr} demonstrate powerful performance on HRNVS but suffer from rendering speed, making them impractical for resource-limited scenarios.
Recent 3D Gaussian Splatting (3DGS) based methods~\cite{shen2025supergaussian,srgs,xie2024supergs,ko2024sequence,3dgsr,xu2025high,yu2024gaussiansr} utilize 3D Gaussian primitives to describe scenes and achieve remarkable acceleration.
However, existing 3DGS methods handle HRNVS at fixed integer scale factors (\eg,$\times 2$ and $\times 4$) and require separate models for different scale factors, ignoring the intrinsic continuous characteristic of 3D world~\cite{hu2024gaussiansr}. They also fail to provide the flexibility for users to adjust rendering accuracy based on available resources.
Therefore, a solution that enables 3D super-resolution of arbitrary scale factors using a single unified model is essential for practical HRNVS.

The most straightforward solution involves directly rendering views of target resolution using a scale-specific 3D model. 
However, this vanilla approach inevitably introduces aliasing artifacts (3DGS in Fig.~\ref{fig:teaser}), indicating its inflexibility and limited generalization ability for finer details.
An alternative solution is to cascade novel view synthesis (NVS) models with an arbitrary-scale super-resolution (SR) upsampler to synthesize novel views of arbitrary scale factors.
Yet, this solution complicates the framework and slows rendering due to the computational overhead of the upsamplers (see Sec.~\ref{sec:efficiency}).

To overcome these limitations, we explore the task of arbitrary-scale 3D Gaussian super-resolution (Arbi-3DGSR), and propose an integrated framework that enables continuous rendering of HR views with arbitrary scale factors (including non-integer ones), using a single 3DGS model.  We observe three challenges in Arbi-3DGSR: anti-aliasing NVS at various scale factors, constraining fine details of HR results without ground truth supervision, and maintaining structural consistency with LR views.  Our framework tackles these challenges through three components: scale-aware rendering, generative prior-guided optimization and progressive super-resolving, respectively.   Specifically, we approach Arbi-3DGSR with the following procedures: (1) \textbf{scale-aware rendering}: we utilize target scale factors to constrain the maximum signal frequency of the 3D model, and adapt the integration window size for accurate pixel shading;(2) \textbf{generative prior-guided optimization}: 2D generative priors from a diffusion model are leveraged to supervise fine details of rendered HR results. To avoid potential inconsistencies introduced by generative priors, we optimize with latent distillation and include orthogonal views for explicit supervision; and (3) \textbf{progressive super-resolving}: the training process is divided into multiple stages, gradually optimizing 3D Gaussian primitives to support larger scale factors while maintaining structural consistency. 

Our contributions are summarized as follows:
\begin{itemize}  

\item  We make the first attempt to explore arbitrary-scale 3D Gaussian super-resolution (Arbi-3DGSR), and propose an integrated framework that enables high-resolution rendering at varying scales using a single model.

\item We introduce three key components: scale-aware rendering, generative prior-guided optimization, and progressive super-resolving to address critical challenges in anti-aliasing rendering, supervising fine details of HR results without ground truth, and maintaining structural consistency across various scale factors, respectively.

\item Extensive experiments on four benchmarks show our superiority in rendering high-quality super-resolved results, including non-integer scale factors (\eg, achieving a PSNR gain of \textbf{6.59 dB} over 3DGS at $\times 5.7$), while maintaining real-time speed of \textbf{85 FPS} for \textbf{1080p} rendering. Detailed ablations and visualizations provide more intuitions.

\end{itemize}

\section{Related Work} 
\subsection{High-Resolution Novel View Synthesis}  
Neural Radiance Field (NeRF) based methods~\cite{cross,supernerf,Lee_2024_CVPR,Huang_2023_CVPR,nerfsr,huang2024assr} enable NVS of arbitrary resolutions with implicit neural representations, but are computationally intensive due to complex architecture designs and volumetric rendering~\cite{srgs}. While 3D Gaussian Splatting (3DGS)~\cite{3dgs} provides efficient rendering of novel views. To provide high-resolution (HR) supervision, existing 3DGS-based methods~\cite{srgs,xie2024supergs,3dgsr, shen2025supergaussian,srgs,xie2024supergs,ko2024sequence} typically generate pseudo HR labels with pre-trained super-resolution models. DLGS~\cite{xu2025high} leverages a dual-lens system to provide geometric hints for HRNVS.  Diffusion-guided methods~\cite{yu2024gaussiansr,li2025leveraging} introduce 2D generative priors and avoid explicit HR supervision. However, the aforementioned methods simply conduct HRNVS of fixed scale factors, ignoring the practical resource constraints and the continuous nature of the 3D world.

\vspace{-5pt}\subsection{Anti-Aliasing 3D Gaussian-Splatting}  
Although 3DGS~\cite{3dgs} enables real-time rendering, it suffers from aliasing artifacts under varying sampling rates (\ie, different focal length and depth). Mip-Splatting~\cite{mipgs} introduces 3D smoothing filter and  2D Mip filter to constrain the maximal sampling frequency of Gaussian primitives. Analytic-Splatting~\cite{liang2025analytic} approximates Gaussian integrals over pixel areas and jointly models transmittance to enhance sampling robustness. SA-GS~\cite{song2024sa} employs a frequency-aware 2D scale-adaptive filter to maintain consistent Gaussian distributions.  MSGS~\cite{yan2024multi} represent scenes with Gaussians of multiple scales, enabling adaptive selection for different sampling conditions to avoid aliasing artifacts.   Unlike the HRNVS task that reconstructs from sparse LR views and renders novel HR views, anti-aliasing 3DGS inherently does not involve resolution changes, and the ground truth at different sampling rates is explicitly provided as supervision.

\vspace{-5pt}\subsection{Arbitrary-Scale Super-Resolution}  
To enable continuous upscaling with a single network, MetaSR~\cite{hu2019meta} dynamically adjusts filter weights based on scale factors.   Zhao~\etal~\cite{zhaotmm} enhance adaptability by fusing local features and scale factors. ArbiSR~\cite{RealArbiSR} employs a dual-level deformable implicit representation to address real-world degradations. COZ~\cite{coz} boost degradation robustness by mixing features and coordinates of multiple points.  BASR~\cite{weng2024best} model scaling with dual degradation representation to achieve cycle consistency. StableSR~\cite{stablesr} introduces progressive aggregation sampling for resolution-agnostic generation.  STAVSR~\cite{stavsr} uses multi-scale priors to distinguish contents across different scales and locations. VideoINR~\cite{chen2022videoinr} employs implicit neural representations to decode videos at arbitrary resolutions and frame rates.  SAVSR~\cite{li2024savsr} introduces omni-dimensional scale-attention and bi-directional fusion to adapt across scales. Inspired by these works, we propose to incorporate scale factors into rendering process to improve the anti-aliasing ability across varying target resolutions.

 \vspace{-4pt}\section{Preliminaries}\label{sec:pre}
We provide a brief review of 3D Gaussian Splatting (3DGS)~\cite{3dgs} and anti-aliasing filters introduced in Mip-Splatting~\cite{mipgs} to achieve aliasing-free novel view synthesis under varying focal lengths and depths. 
\vspace{-4pt}\subsection{3D Gaussian Splatting} 3DGS models a scene with a set of explicit points $\{\boldsymbol{p}_i\}_{i=1}^N$ parameterized by opacity $\boldsymbol{\alpha}$, color $\boldsymbol{c}$, and 3D Gaussian primitive based geometry $\{\boldsymbol{G}_i\}_{i=1}^N$.
 Each Gaussian primitive is described with a full 3D covariance matrix $\boldsymbol{\Sigma} \in \mathbb{R}^{3 \times 3}$ and center position $\boldsymbol{\mu}  \in \mathbb{R}^{3 \times 1}$ as follows:
 \vspace{-2pt}\begin{equation} \label{eq:3dgs}
G^{3D}(\boldsymbol{x})=e^{ -\frac{1}{2}(\boldsymbol{x}-\boldsymbol{\mu})^{\top} \boldsymbol{\Sigma}^{-1}(\boldsymbol{x}-\boldsymbol{\mu})},
\end{equation}
where $\boldsymbol{\Sigma}$ is defined with a scaling matrix $\boldsymbol{S} \in \mathbb{R}^{3 \times 1}$ and rotation matrix $\boldsymbol{R} \in \mathbb{R}^{3 \times 3}$ as $\boldsymbol{R}\boldsymbol{S}\boldsymbol{S}^\top\boldsymbol{R}^\top$.

\vspace{-6pt}\subsection{Anti-Aliasing Filtering}
Aliasing occurs when a scene is reconstructed with fixed focal length and depth but rendered with lower sampling rates. Mip-Splatting~\cite{mipgs} band-limits each Gaussian primitive through a two-stage filtering, such that its highest signal frequency remains below the Nyquist frequency determined by input views.

\noindent \textbf{3D Smoothing Filter.}  Projecting a continuous 3D signal onto the 2D screen plane is a sampling operation. According to Nyquist–Shannon sampling theorem~\cite{nyquist1928certain,shannon1949communication}, given discrete samples taken at frequency $\hat{r}$, reconstructed continuous signals can theoretically yield frequency $r$ up to $\frac{\hat{r}}{2}$ (\ie, $\hat{r} \geq 2r$). Thus, considering a Gaussian primitive $G_i^{3D}$ observed in $K$ views with focal lengths $\{f_k\}_{k=1}^K$ in pixel units and depths $\{d_k\}_{k=1}^K$ in 3D world space, its maximum sampling rate is:
\vspace{-4pt}\begin{equation}\label{eq:max_freq} 
\resizebox{0.6\linewidth}{!}{$
\hat{r}_i= \max \left(\left\{\mathbb{I}_k\left(G^{3D}_i\right) \cdot \frac{f_k}{d_k}\right\}_{k=1}^K\right),  \vspace{-6pt}
$}
\end{equation}
where $\mathbb{I}_k (G^{3D}_i)$ evaluates the visibility of $G^{3D}_i$. To cap the highest frequency of  reconstructed 3D model, Mip-Splatting imposes a low-pass Gaussian filter on each 3D Gaussian primitive $G^{3D}_i$, with hypermeter $\gamma$ controlling the filter size:
  \vspace{-2pt}\begin{equation}\label{eq:3dmipfilter}    
 \resizebox{0.43\textwidth}{!}{$ 
G_i^{3D}(\boldsymbol{x})_{mip}
=\sqrt{\frac{\left|\boldsymbol{\Sigma}_i\right|}{\left|\boldsymbol{\Sigma}_i+\frac{\gamma}{\hat{r}_i} \cdot \mathbf{I}\right|}} e^{-\frac{1}{2}\left(\boldsymbol{x}-\boldsymbol{\mu}_i\right)^\top\left(\boldsymbol{\Sigma}_i+\frac{\gamma}{\hat{r}_i} \cdot \mathbf{I}\right)^{-1}\left(\boldsymbol{x}-\boldsymbol{\mu}_i\right)}.  \vspace{-2pt}
$}
\end{equation}

\noindent \textbf{2D Mip Filter.}
During rendering, each 3D Gaussian in world space is projected to image plane as a 2D Gaussian, characterized by position $\hat{\boldsymbol{\mu}}=\boldsymbol{P}\boldsymbol{W}[\boldsymbol{\mu}, 1]^{\top}$ and covariance $\hat{\boldsymbol{\Sigma}}=\boldsymbol{J} \boldsymbol{W} \boldsymbol{\Sigma} \boldsymbol{W}^\top \boldsymbol{J}^\top$. Here $\boldsymbol{P}$, $\boldsymbol{W}$ and $\boldsymbol{J}$ denote the projection matrix,  extrinsic matrix, and Jacobian of projective transformation, respectively.   When shading a pixel $\hat{\boldsymbol{x}}$,  Mip-Splatting imposes a 2D Gaussian filter over 2D Gaussians, thereby providing an integration window area and effectively preventing  2D Gaussian signals from being too small:
\begin{equation}\label{eq:2dmip}  \vspace{-4pt}
\resizebox{0.43\textwidth}{!}{$
G_i^{2D}(\hat{\boldsymbol{x}})_{mip} = \sqrt{\frac{\left|\hat{\boldsymbol{\Sigma}}_i\right|}{\left|\hat{\boldsymbol{\Sigma}}_i+\varepsilon \cdot \mathbf{I}\right|}} e^{-\frac{1}{2}\left(\hat{\boldsymbol{x}}-\hat{\boldsymbol{\mu}}_i\right)^\top\left(\hat{\boldsymbol{\Sigma}}_i+\varepsilon\cdot \mathbf{I}\right)^{-1}\left(\hat{\boldsymbol{x}}-\hat{\boldsymbol{\mu}}_i\right)},
$}
\end{equation} 
where $\varepsilon$ is a hypermeter to cover a single pixel. The color of pixel $\hat{\boldsymbol{x}}$ is then accumulated as follows:
\begin{equation}   
\resizebox{0.91\linewidth}{!}{$
\boldsymbol{C}(\hat{\boldsymbol{x}})=\sum_{i = 1}^{N}  \alpha_i c_i G_i^{2D}(\hat{\boldsymbol{x}})_{mip} \prod_{j=1}^{i-1}\left(1-\alpha_j G_j^{2D}(\hat{\boldsymbol{x}})_{mip} \right), 
$}
\end{equation}
where $\alpha_i$ and $c_i$ are the opacity and color of the $i$-th primitive.

\section{Method}\label{sec:method}
Given a set of low-resolution (LR) views, our goal is to reconstruct a 3D scene and render high-resolution (HR) views at a target scale factor $s$ (\ie, the enlarging ratio from LR to HR views). We introduce a unified framework that comprises three key components: scale-aware rendering, generative prior-guided optimization, and progressive super-resolving. Scale-aware rendering is applied in both training and inference, enabling aliasing-free HRNVS at target scale factor $s$. During training, generative prior-guided optimization constrains fine details in the absence of ground truth, while progressive super-resolving gradually increases target scales to preserve structural consistency across varying scale factors. 
 
\vspace{-4pt} \subsection{Scale-Aware Rendering}\label{sec:sar}
The highest frequency of the reconstructed 3D Gaussian signal is fixed, whereas changing output resolution alters image-plane sampling density.  A lower sampling rate can fall below the Nyquist frequency and cause aliasing, while a higher rate potentially blurs detail. Adjusting Gaussian bandwidth and per-pixel integration window both help to align the Gaussian signal with the pixel area. Therefore, we propose the following scale-aware filterings.

\noindent \textbf{3D Scale-Aware Smoothing Filter.}
 The maximum sampling rate of a Gaussian primitive is jointly determined by pixel density $\rho $, focal length $f$, and camera depth $d$ with $\hat{r}=\frac{f\cdot \rho}{d}$. Hence, $\hat{r} (s) =\frac{f\cdot \rho \cdot s}{d}$ when the output resolution is scalable with factor $s$. For a Gaussian primitive $G_i^{3D}$ that is visible in $K$ views, its maximum sampling rate is:
\vspace{-4pt}\begin{equation}\label{eq:max_freq_ours} 
\hat{r}_i (s)=\max \left(\left\{\mathbb{I}_k\left(G^{3D}_i\right) \cdot \frac{f_k \cdot s_k }{d_k} \right\}_{k=1}^K \right), \vspace{-2pt}
\end{equation} 
where $s_k$ is the scale factor of the $k$-th camera. $\rho$ is folded into hyperparameter $\gamma$ when constraining the highest signal frequency using Eq.~\ref{eq:3dmipfilter}. By integrating the scale factor, our method ensures a more adaptive and accurate Gaussian bandwidth constraint across varying resolutions.

\begin{figure}
    \centering
  \includegraphics[width=1.02\linewidth,clip=true, trim=6pt 10pt 8pt 4pt]{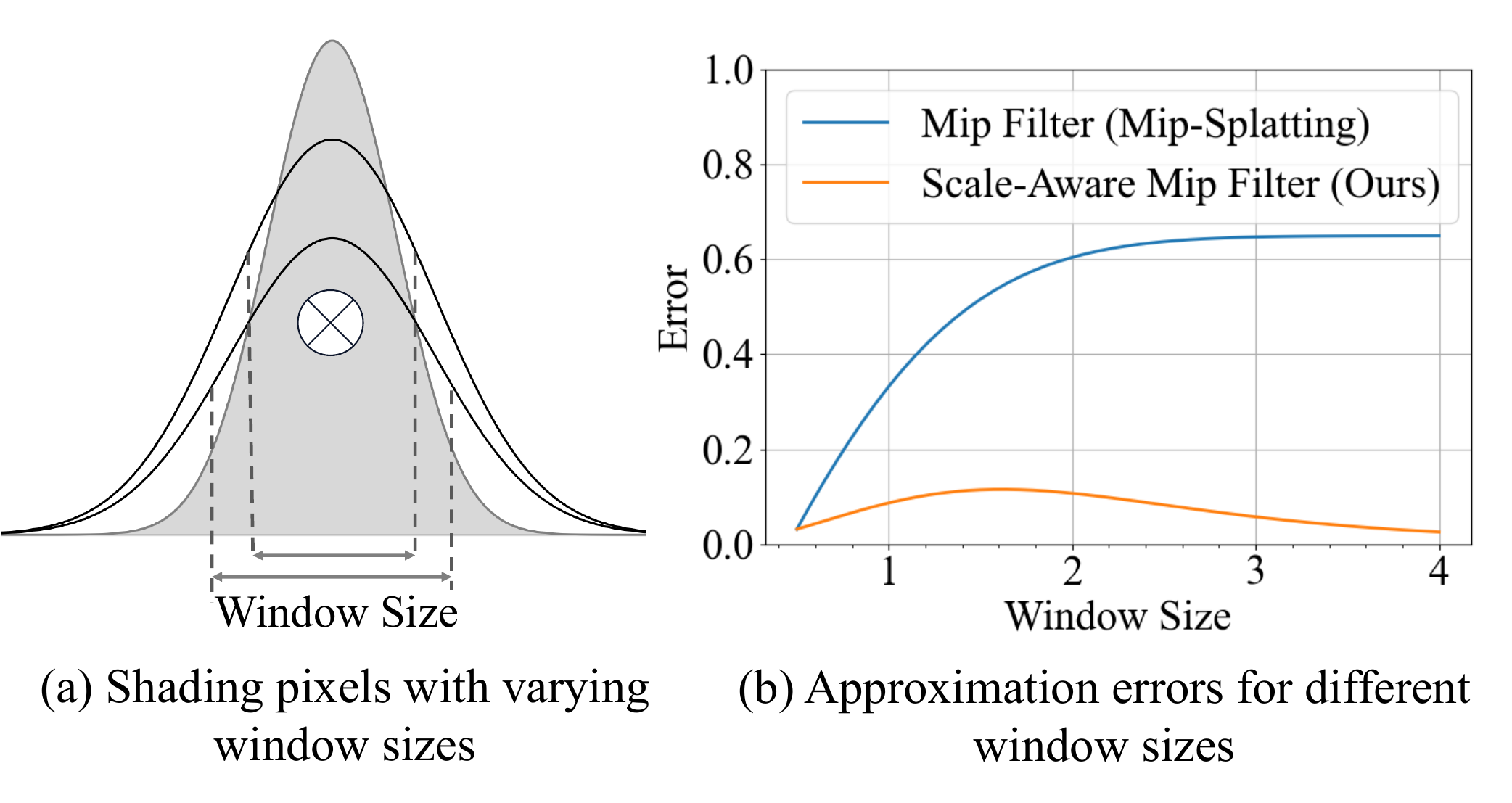}\vspace{-6pt}
    \caption{(a) Accurate pixel shading requires aligning the integration window with the actual pixel size. (b) Approximation error analysis regarding the window size, where the proposed method results in low approximation errors.} \vspace{-0.11in}
    \label{fig:mip2d}
\end{figure}

\noindent \textbf{2D Scale-Aware Mip Filter.}  
When shading pixels with the Gaussian signal, the integration window size depends on the number of pixels (see Fig.~\ref{fig:mip2d}(a)), with more pixels implying a smaller window. To study the effect of window size on integration, we analyze the 1D approximation error by comparing the CDF of a standard Gaussian with that of the vanilla 2D Mip filter. As shown by the blue curve in Fig.~\ref{fig:mip2d}(b), as the actual window size increases, integrating with a fixed window area accumulates approximation error. This reveals the limitation of static integration for scale-dependent signal capture. Therefore, we adaptively set the window size by choosing the 2D Gaussian filter variance as $\varepsilon_k=\frac{\varepsilon}{s_k}$. The scale-aware 2D Gaussian filter is then incorporated into 2D Gaussian signals with:
\begin{equation} 
\resizebox{0.43\textwidth}{!}{$
G_i^{2D}(\hat{\boldsymbol{x}})_{mip} = \sqrt{\frac{\left|\hat{\boldsymbol{\Sigma}}_i\right|}{\left|\hat{\boldsymbol{\Sigma}}_i+\varepsilon_k \cdot \mathbf{I}\right|}} e^{-\frac{1}{2}\left(\hat{\boldsymbol{x}}-\hat{\boldsymbol{\mu}}_i\right)^\top\left(\hat{\boldsymbol{\Sigma}}_i+\varepsilon_k\cdot \mathbf{I}\right)^{-1}\left(\hat{\boldsymbol{x}}-\hat{\boldsymbol{\mu}}_i\right)}.
$}
\end{equation}
 The orange curve in Fig.~\ref{fig:mip2d}(b) shows that the proposed scale-aware Mip filter consistently keeps a lower approximation error compared to the vanilla Mip filter. This highlights its effectiveness in precisely capturing signals (see Sec.~\ref{sec:ablation}).

\vspace{-4pt}\subsection{Generative Prior-guided Optimization} 
For Arbi-3DGSR, only input LR views are available.
To constrain fine details in the rendered HR views, we use generative priors from StableSR~\cite{stablesr} as 
texture-rich references. However, directly optimizing with generated reference causes view inconsistencies due to the diverse and stochastic nature of the generation task (see Sec.~\ref{sec:ablation}).
To this end, we propose the generative latent distillation and optimize with orthogonal reference refinement, ensuring both high perceptual quality and structural coherence across views.
\begin{figure}
\centering
      \includegraphics[width=0.98\linewidth]{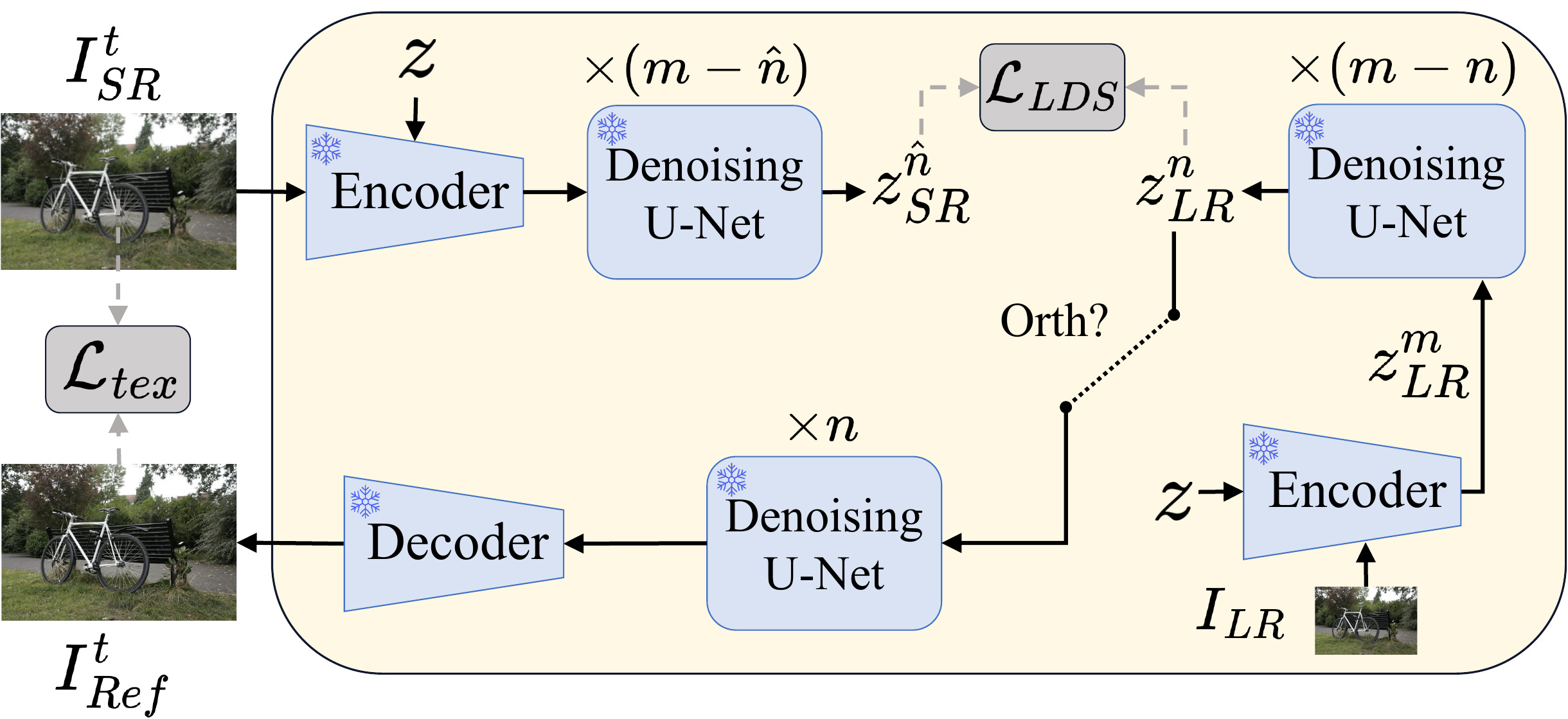}\vspace{-6pt}
        \caption{Generative prior-guided optimization, where generative priors are leveraged to constrain details in rendered HR views. To alleviate view inconsistency introduced by generative priors, optimization is conducted in the latent space, and texture supervision is applied only to orthogonal views. }\vspace{-4pt}
\label{fig:diffusion}
\end{figure}

\noindent \textbf{Generative Latent Distillation.}
In Fig.~\ref{fig:diffusion}, given an LR view $I_{LR}$, we sample a random Gaussian noise $z$, and embed it into the latent space conditioned on $I_{LR}$.
This results in a noisy latent $z^m_{LR}$, where $m$ is the initial denoising timestep.
$z^m_{LR}$ is then iteratively denoised through a UNet-based diffusion process $z^{m-1}_{LR}=UNet(z^m_{LR}, m)$, progressively producing an intermediate latent $z^n_{LR}$ that is enriched with structural information.   Meanwhile, the scale-aware rendered view $I^t_{SR}$ at $t$-th training stage undergoes the same conditional diffusion process. This yields an intermediate latent $z^{\hat{n}}_{SR}$, where $\hat{n}$ is an early denoising timestep ($\hat{n} >n$). $z^n_{LR}$ therefore contains richer structural information from generative priors.  To inject structural information from $z^n_{LR}$ into $z^{\hat{n}}_{SR}$ and guide the optimization, we introduce the Latent Distillation Sampling (LDS) loss $\mathcal{L}_{LDS}$ that computes the gradient as follows:
\begin{equation} \label{eq:lds}
\resizebox{0.43\textwidth}{!}{$
\begin{aligned}
\nabla_\theta \mathcal{L}_{\mathrm{LDS}}(\theta) =  \mathbb{E}_{\hat{n}}\left[w(\hat{n}) \cdot \left (\epsilon_\phi\left(z^{\hat{n}}_{SR} ; I^t_{SR}, \hat{n}\right) -\epsilon_\phi\left(z^n_{LR};  I_{LR}, n\right)\right) \frac{\partial I^t_{SR}}{\partial \theta}\right],  
\end{aligned}
$}
\end{equation}
where $\epsilon_\phi(\cdot)$ indicates predicting noise with UNet.  $w(\hat{n})$ is a timestep $\hat{n}$ related weighting function. $\theta$ denotes learnable parameters of 3D Gaussian primitives. Unlike the Score Distillation Sampling loss~\cite{pooledreamfusion} that compares predicted noise with attached Gaussian noise of the same timestep, Eq.~\ref{eq:lds} minimizes the noise discrepancy between asynchronous latents $z^{\hat{n}}_{SR}$ and $z^n_{LR}$.  This encourages $z^{\hat{n}}_{SR}$ to approximate the detailed texture in $z^n_{LR}$, while fully leveraging the structural information of LR views. Furthermore, minimizing the noise discrepancy instead of pixel discrepancy effectively provides structural supervision while tolerating pixel misalignment introduced by generative priors. We provide further analysis and theoretical derivation of LDS loss in Sec.~\ref{sec:ablation} and supplementary materials, respectively.

\noindent \textbf{Orthogonal Reference Refinement.} 
While generative latent distillation provides inconsistency-tolerant supervision, explicit pixel-wise supervision is crucial for enhancing fine-grained texture details. To address potential inconsistencies between adjacent views, we employ an orthogonal view strategy: for each scene, we identify a subset of views that are approximately orthogonal (\ie, the number of views is scene-dependent) for pixel-wise refinement. As shown in Fig.~\ref{fig:diffusion}, for orthogonal views, the latent variable $z^n_{LR}$ is further denoised and decoded to obtain an HR reference $I^t_{Ref}$.  The texture loss  $\mathcal{L}_{tex}$ is conducted with Eq.~\ref{eq:loss_tex} below:
\begin{equation} \label{eq:loss_tex}
\mathcal{L}_{tex}=\mathbb{I}_{ortho} \cdot \| I^t_{SR} - I^t_{Ref} \|^2,
\end{equation}               
where $\mathbb{I}_{ortho}$ indicates whether current view is orthogonal. The orthogonal reference refinement encourages learning from non-overlapping views, avoiding conflicting information and preserving geometric consistency.

\begin{figure}
    \centering
    \vspace{-2pt}\includegraphics[width=0.9\linewidth]{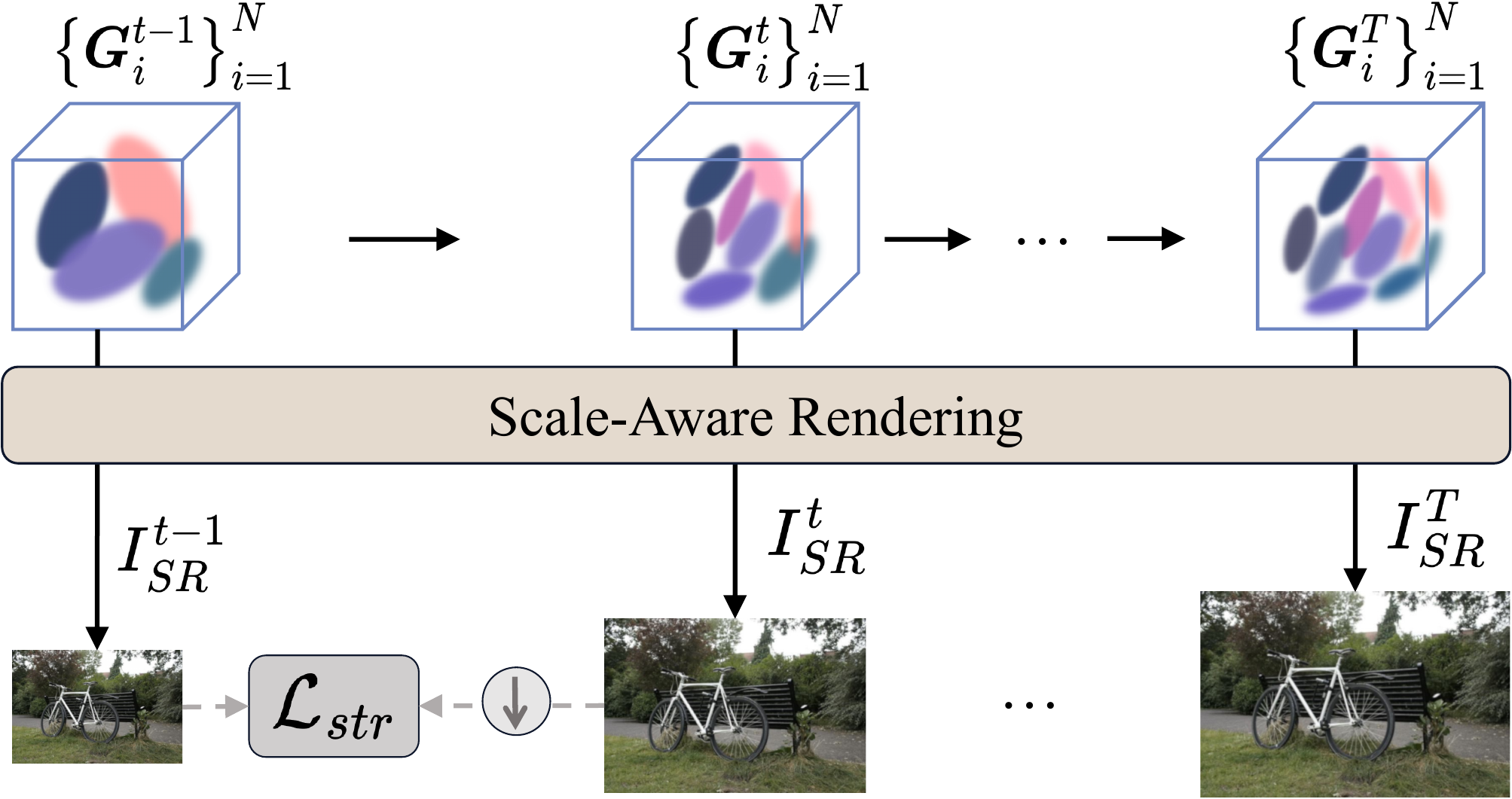}\vspace{-6pt}
    \caption{Progressive super-resolving. The training process is divided into multiple stages, with each stage following the same mechanism while progressively rendering higher-resolution views. Structural loss is applied between adjacent stages to ensure consistency across scale factors.}  \vspace{-0.1in}
    \label{fig:progressive}
\end{figure}

\begin{table*}[t]
\centering
\Large
\resizebox{1\linewidth}{!}{%
 {\renewcommand{\arraystretch}{1.1}
\begin{tabular}{c|c|ccc|ccc|ccc|ccc|ccc}
\toprule
 \multicolumn{2}{c|}{\multirow{2}{*}{Method}} & \multicolumn{3}{c|}{$\times 2$} & \multicolumn{3}{c|}{$\times 4$}  &  \multicolumn{3}{c|}{$\times 8$} &   \multicolumn{3}{c|}{$\times 3.5$} &   \multicolumn{3}{c}{$\times 5.7$} \\
\cmidrule{3-17} 
  \multicolumn{2}{c|}{}  &   PSNR$\uparrow$  &   SSIM$\uparrow$  &  FID$\downarrow$   
   &  PSNR$\uparrow$  &   SSIM$\uparrow$  &  FID$\downarrow$   
   & PSNR$\uparrow$  &   SSIM$\uparrow$  &  FID$\downarrow$   
    & PSNR$\uparrow$  &   SSIM$\uparrow$  &  FID$\downarrow$   
    & PSNR$\uparrow$  &   SSIM$\uparrow$  &  FID$\downarrow$   \\
\midrule 
 \multirow{8}{*}{ \rotatebox{90}{Blender}} & 3DGS  & 19.66 &  0.834 & 129.567 & 17.84 &  0.789 &  208.166 & 16.86 & 0.797 & 244.212 & 18.10 & 0.792 &  198.147 & 17.28 & 0.790 & 229.619 \\
\cmidrule{2-17}
 & Bicubic & 20.36 & 0.853 & 132.513 &  19.67 &  0.831 & 178.170  & 19.24 &  0.831 & 182.420  & 19.76 & 0.833 & 176.546 & 19.46 & 0.829 & 181.065 \\
 & ArbiSR & 19.34 & 0.827 & 152.266& 18.23 & 0.803 & 185.134 & 17.98 & 0.817 & 174.873	& 18.33 & 0.803 & 185.430 & 18.08 & 0.808 & 179.436 \\	
 & StableSR & 18.63 & 0.788 & 120.385 & 18.25 & 0.785 & \cellcolor{yellow!30}95.086 & 17.98 & 0.797 & \cellcolor{yellow!30}94.957	 & 18.31 & 0.785 & \cellcolor{yellow!30}95.720 & 18.12 & 0.790 & \cellcolor{yellow!30}94.853 \\
\cmidrule{2-17}
& Mip-Splatting & 23.33 & 0.896 & \cellcolor{yellow!30}78.907 & 22.25 & 0.870 & 109.435 & 21.57 & \cellcolor{yellow!30}0.858 & 113.124 & 22.40 & 0.874 & 108.101 & 21.92 & \cellcolor{yellow!30}0.862 & 111.875 \\
& Analytic-Splatting & \cellcolor{yellow!30}25.23 &	\cellcolor{yellow!30}0.915	&	103.452	&	\cellcolor{yellow!30}23.57	& \cellcolor{yellow!30}0.873	&	141.302	& \cellcolor{yellow!30}22.51 & 0.847	&	146.244 & \cellcolor{yellow!30}23.82	& \cellcolor{yellow!30}0.881	&139.486	& \cellcolor{yellow!30}23.02 &	0.857	& 144.521 \\
\cmidrule{2-17}
& GaussianSR  &23.25	 & 0.885	 & 94.017	 & 23.03	 & 0.868 &	118.021 & 	22.37	 & 0.856	 &	117.618 & 23.13	 & 0.871 &	117.458	 & 22.73	 & 0.861	 & 118.465\\
 & Ours &  \cellcolor{red!30}25.60	 & \cellcolor{red!30}0.925	 & 	\cellcolor{red!30}66.515	 & \cellcolor{red!30}24.32	 & \cellcolor{red!30}0.899	 & \cellcolor{red!30}86.270	 & \cellcolor{red!30}23.40	 & \cellcolor{red!30}0.879	 & \cellcolor{red!30}87.984 & \cellcolor{red!30}24.53 & 	\cellcolor{red!30}0.904	  & 	\cellcolor{red!30}85.788	 & \cellcolor{red!30}23.87 & 	\cellcolor{red!30}0.888	 & \cellcolor{red!30}87.429 \\
 \toprule  
 \addlinespace[2pt]
\toprule
 \multirow{8}{*}{\rotatebox{90}{Mip-NeRF360}} & 3DGS & 23.55 & 0.710 & 49.057 & 20.98 & 0.610 & 68.864	& 19.92 & 0.632 & 85.628 & 21.32 & 0.619 &64.968 & 20.33 & 0.609 & 77.907 \\  
\cmidrule{2-17}
& Bicubic & 25.92 & 0.738 &  44.731 &  25.09 &  0.687 & 47.441 &  \cellcolor{red!30}24.86 & \cellcolor{red!30}0.730 &  47.551 & 25.20 &  0.689 &  46.805 & 24.96 & 0.699 & 47.450 \\ 	
 & ArbiSR & 25.14 & 0.727 & 56.838  & 23.92 & 0.671 & 57.471  & 23.87 & 0.714 & 54.983 & 24.02 & 0.674 & 57.714 & 23.86 & 0.683 & 56.061 \\ 
 & StableSR & 20.85 & 0.480 & 107.550 & 20.54 & 0.507 & 106.957 & 20.45 & 0.605 & 106.950 & 20.58 & 0.495 & 107.272 & 20.49 & 0.551 & 106.968 \\		
\cmidrule{2-17}
& Mip-Splatting &  \cellcolor{yellow!30}26.10 & 0.750& \cellcolor{yellow!30}35.931& \cellcolor{yellow!30}25.17& \cellcolor{yellow!30}0.696& 38.950& 24.51& 0.718&  41.025& \cellcolor{yellow!30}25.30& \cellcolor{yellow!30}0.700& 38.294& \cellcolor{red!30}25.02& \cellcolor{yellow!30}0.702& 39.089 \\
& Analytic-Splatting & 25.80 & \cellcolor{yellow!30}0.754 & \cellcolor{red!30}27.536 & 24.02 & 0.652 & \cellcolor{red!30}28.421 & 23.04 & 0.615 & \cellcolor{red!30}28.500 & 24.27 & 0.668 & \cellcolor{red!30}28.180 &23.41 & 0.621 & \cellcolor{red!30}28.734 \\
\cmidrule{2-17}
& GaussianSR & 24.93	 &  0.660	 & 	103.160	 &  24.33	 &  0.650	 & 	105.013	 &  24.10	 &  0.710	 & 	104.943 & 	24.42	 &  0.646	 & 	104.548  &  24.20	 &  0.674	 & 	105.031 \\
 & Ours & \cellcolor{red!30}26.23	& \cellcolor{red!30}0.764	& 	36.524	& \cellcolor{red!30}25.18& \cellcolor{red!30}	0.703& \cellcolor{yellow!30}38.526	& 	\cellcolor{yellow!30}24.85& 	\cellcolor{yellow!30}0.725&  \cellcolor{yellow!30}38.645 & \cellcolor{red!30}25.32	&  \cellcolor{red!30}0.709&  \cellcolor{yellow!30}37.924		& \cellcolor{yellow!30}24.99	& \cellcolor{red!30}0.704	& \cellcolor{yellow!30}38.606 \\
 \toprule  
 \addlinespace[2pt]
\toprule
  \multirow{8}{*}{\rotatebox{90}{Tanks\&Temples}} & 3DGS  & 19.45 & 0.711 &  72.678 & 16.24 & 0.516 & 135.005 & 14.86 & 0.468 &176.733 & 16.67 & 0.541 & 121.909 & 15.43 & 0.479 & 161.616 \\ 		
\cmidrule{2-17}
& Bicubic & 21.80 & 0.742 &79.308& 20.19 &  0.599 &127.444	 &  19.27 & 0.554 &141.343	&  20.45 &  0.619 & 	120.743 &  19.74 & 0.567 &137.071 \\ 	
& ArbiSR & 20.85 & 0.719 & 148.429 & 18.90 & 0.581 &188.368	& 18.40 & 0.547 & 165.008 & 19.11 & 0.598 & 188.735	& 18.66 & 0.557 & 176.148	\\ 		
& StableSR & 19.41 & 0.653 &90.379 & 18.54 & 0.538 & 86.072 & 17.82 & 0.498 & \cellcolor{red!30}85.640& 18.69 & 0.553 &  86.732 & 18.19 & 0.510 & \cellcolor{red!30}84.504 \\		
\cmidrule{2-17}
& Mip-Splatting & \cellcolor{yellow!30}22.78 	 &  \cellcolor{yellow!30}0.806	 & 	\cellcolor{yellow!30}53.183 & 	\cellcolor{yellow!30}20.97	 & \cellcolor{yellow!30}0.667	 & \cellcolor{yellow!30}86.023 & 	\cellcolor{yellow!30}19.90	 & \cellcolor{yellow!30}0.592	 & 99.700 & \cellcolor{yellow!30}21.27	 & \cellcolor{yellow!30}0.690  & 	\cellcolor{yellow!30}79.479	 & \cellcolor{yellow!30}20.44	 & \cellcolor{yellow!30}0.621	 & 	95.086 \\
& Analytic-Splatting & 21.75 & 0.754 &  73.966 & 19.42 & 0.577 & 127.344 & 18.20 & 0.498 &  143.717 &  19.78 & 0.605 &  118.811 & 18.75 & 0.525 &  137.808 \\
\cmidrule{2-17}
& GaussianSR &  21.90 & 	0.738 & 80.779	&  20.63	& 0.623	& 	102.969	& 19.60	&  0.560	& 	112.046	&  20.91	& 0.643	& 	97.359	& 20.13	& 0.584	& 	109.541 \\
 & Ours & \cellcolor{red!30}22.94	& \cellcolor{red!30}0.821	& \cellcolor{red!30}48.659	& \cellcolor{red!30}21.14	& \cellcolor{red!30}0.686	&	\cellcolor{red!30}75.591	& \cellcolor{red!30}19.99 &	\cellcolor{red!30}0.600	&	\cellcolor{yellow!30}89.928	& \cellcolor{red!30}21.47	& \cellcolor{red!30}0.710	& 	\cellcolor{red!30}69.519	& \cellcolor{red!30}20.57	& \cellcolor{red!30}0.635	&	\cellcolor{yellow!30}85.057\\
 \toprule  
 \addlinespace[2pt]
\toprule
  \multirow{8}{*}{\rotatebox{90}{Deep Blending}} & 3DGS  & 25.57 & 0.794 & 159.958 & 23.70 & 0.736 & 222.691 & 22.81 & 0.748 & 258.275 & 23.99 & \cellcolor{yellow!30}0.741 & 211.857 & 23.19 & 0.738 & 243.801 \\		
\cmidrule{2-17} 
& Bicubic & 26.40 & 0.817 & 140.654 &  25.60 &  0.793 & 173.301 & 25.27 & 0.814 & 185.594	& 25.75 & 0.793 & 169.449 &  25.46 &  0.802 & 180.383 \\		
& ArbiSR & 26.09 & 0.817 & 161.001 & 24.87 & 0.787 &  195.876 & 24.62 & 0.808 & 200.379 & 25.03 & 0.787 & 192.530 & 24.74 & 0.797 & 199.635 \\		
& StableSR & 24.31 & 0.755 &  \cellcolor{yellow!30}131.813 & 23.81 & 0.729 & \cellcolor{red!30}127.640 & 23.57 & 0.749 & \cellcolor{red!30}128.898 & 23.90 & 0.729 & \cellcolor{red!30}129.270 & 23.70 & 0.736 & \cellcolor{red!30}128.558\\	 \cmidrule{2-17}
& Mip-Splatting & 26.75  &  0.830	& 137.672	& 25.90	&  0.805	& 174.613	& \cellcolor{yellow!30}25.52	& \cellcolor{yellow!30}0.818	& 189.308 & 26.06	&  0.806	& 168.978 &	25.73	& 0.810	& 182.643 \\	
& Analytic-Splatting & 26.26 & 0.805	 & 159.607 & 25.20 & 0.755	& 	201.319 & 24.70 & 0.757	& 217.368 & 25.39 & 0.761	& 195.165 & 24.95 & 0.752	& 	210.790 \\
\cmidrule{2-17}
& GaussianSR & \cellcolor{yellow!30}26.89	& \cellcolor{yellow!30}0.843	&	134.990	& \cellcolor{yellow!30}26.30	&\cellcolor{yellow!30}0.826	&163.499	&25.91	& 0.832	&	172.887	& \cellcolor{yellow!30}26.45	& \cellcolor{yellow!30}0.827	& 159.877	& \cellcolor{yellow!30}26.13	&\cellcolor{yellow!30}0.829&169.988\\
 & Ours & \cellcolor{red!30}27.44 &  \cellcolor{red!30}0.861	& \cellcolor{red!30}106.533 &  \cellcolor{red!30}26.57 &  \cellcolor{red!30}0.835	& \cellcolor{yellow!30}140.633 &  \cellcolor{red!30}26.14 &  \cellcolor{red!30}0.836	& 	\cellcolor{yellow!30}154.810 & \cellcolor{red!30}26.75 &  \cellcolor{red!30}0.837	& 	\cellcolor{yellow!30}134.836 &  \cellcolor{red!30}26.38 &  \cellcolor{red!30}0.835	& \cellcolor{yellow!30}148.951 \\
\bottomrule
\end{tabular}  }
}  
\caption{Quantitative results on Blender, Mip-NeRF360, Tanks\&Temples, and Deep Blending dataset, with the best and second-best results highlighted in \colorbox{red!30}{red} and \colorbox{yellow!30}{yellow}, respectively. The proposed method achieves the overall best performance, demonstrating its effectiveness in rendering high-fidelity results and preserving consistency across different scale factors. 
} \vspace{-0.1in}
\label{tab:quantitative}
\end{table*}

\vspace{-4pt}\subsection{Progressive Super-Resolving} 
As we use a single 3D model for arbitrary-scale HR rendering, each output must remain structurally consistent with LR views. Otherwise, even slight misalignments magnify into visible warping or registration errors. To this end, training is divided into multiple stages to progressively accommodate various scale factors (see Fig.~\ref{fig:progressive}). Each stage adopts the same working mechanism and loss function, with the maximum scale factor increasing at each stage, enabling the model to gradually refine details at higher resolutions.  
At the $t$-th stage, the 3D Gaussian primitives are initialized from those learned in the previous $(t-1)$-th stage.  For each HR view, the rendering scale factor is randomly selected from  $\{s^1, s^2, \ldots, s^t\}$, where $s^t$ is the largest scale factor used at the $t$-th stage.  To maintain structural consistency across varying scale factors, a structure loss is imposed between current $t$-th stage HR view and corresponding lower-resolution view from previous ($t{-}1$)-th stage.  Specifically, given the current scale factor $s^i$, the previous stage renders with $s^{i-1}$ to ensure a smooth transition between scales. The structure loss $\mathcal{L}_{str}$ is conducted between the rendered outputs $I^t_{SR}$ and $I^{t-1}_{SR}$ as follows:
\vspace{-2pt}\begin{align}\label{eq:struct}
\mathcal{L}_{str} 
&= (1 - \lambda) \, \mathcal{L}_{\mathrm{MSE}}\left(
    \mathcal{D}\left(I^t_{SR}, \frac{s^i}{s^{i-1}}\right), I^{t-1}_{SR}
\right) \nonumber \\
&\quad + \lambda \, \mathcal{L}_{\mathrm{D\text{-}SSIM}}\left(
    \mathcal{D}\left(I^t_{SR}, \frac{s^i}{s^{i-1}}\right), I^{t-1}_{SR}
\right)
\end{align}\vspace{-2pt}
where $\mathcal{D}(I^t_{SR},\frac{s^i}{s^{i-1}})$ indicates downsampling the HR  $I^t_{SR}$ by $\frac{s^i}{s^{i-1}}$. $\lambda$ is the hyperparameter controlling the balance between each loss term.
Applying structural loss between adjacent stages helps preserve structural similarity, ensuring smooth transitions and continuity across different scales. 
As a result, our overall objective function is described below: 
\begin{equation} 
\mathcal{L}=\lambda_1\mathcal{L}_{LDS}+\lambda_2\mathcal{L}_{tex}+\lambda_3\mathcal{L}_{str},
\end{equation}
where $\lambda_1$, $\lambda_2$ and $\lambda_3$ denote weights of each loss term.

 \begin{figure*}[t]
    \centering
\includegraphics[width=1\linewidth,clip=true, trim=26pt 20pt 2pt  0 ]{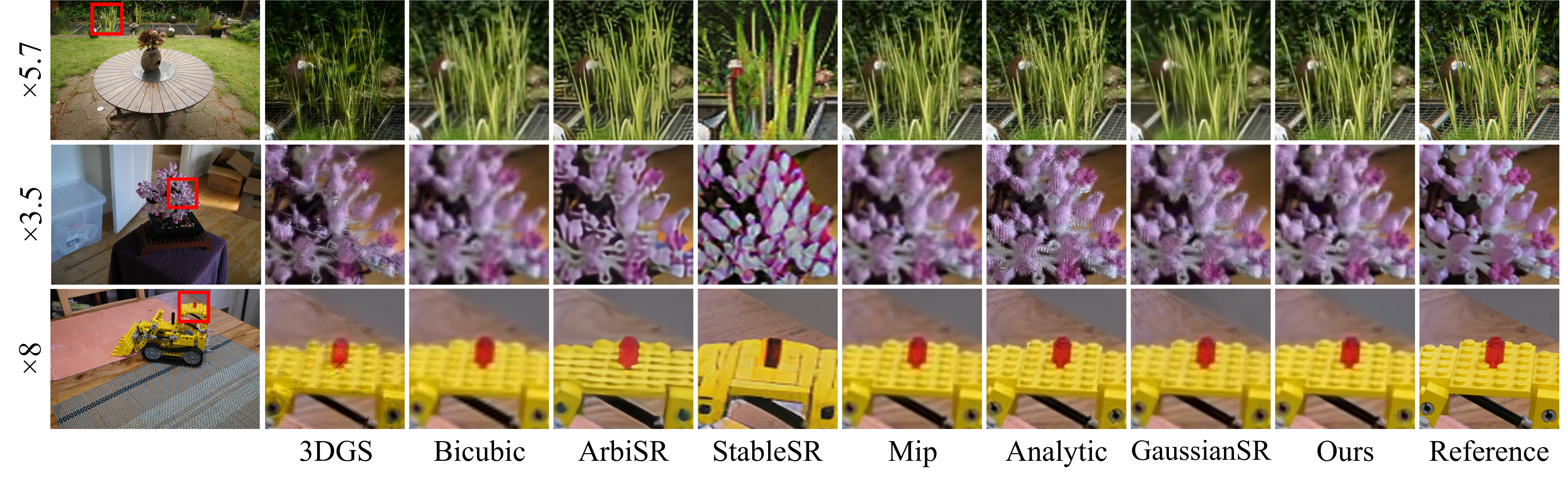}\vspace{-0.1in}
     \caption{Qualitative comparisons on the Mip-NeRF360 dataset, where Mip and Analytic denote Mip-Splatting and Analytic-Splatting, respectively. Please zoom in for better results. 
       As can be seen, 3DGS contains aliasing artifacts (\eg, 1st column).   StableSR changes the contents of the rendered view (\eg, lego in the 3rd row). Analytic-Splatting and GaussianSR generate high-frequency artifacts.  In contrast, the proposed method effectively renders high-fidelity results with rich details. 
     }\vspace{-4pt}
    \label{fig:mip360}
\end{figure*}

\section{Experiments}\label{sec:exp} 
\subsection{Experimental Settings}  
\noindent \textbf{Datasets.}  
Experiments are conducted on four benchmark datasets: Blender~\cite{nerf}, Mip-NeRF360~\cite{mipnerf360}, Tanks\&Temples~\cite{tandt}, and Deep Blending~\cite{db}, following the same train-test splits as 3DGS~\cite{3dgs}.  LR input views are generated by downsampling the original full-resolution images by a factor of 8 using Bicubic interpolation. $\times 2$, $\times 4$ and $\times 8$ are adopted as the upscale factors across three training stages. Notably, original full-resolution images are not used during training, as ground-truth is unavailable in the super-resolution scenario.  To assess generalization ability across arbitrary scales, we evaluate both integer ($\times 2$, $\times 4$ and $\times 8$) and non-integer ($\times 3.5$ and $\times 5.7$) scale factors, with corresponding ground truth views obtained by downsampling the original HR images to the target resolution.

\noindent \textbf{Metrics.}
PSNR and SSIM are employed as error-based metrics to evaluate the performance.
FID~\cite{heusel2017gans} is adopted to assess the perceptual quality.  The original full-resolution views serve as the reference distribution.

\noindent \textbf{Baselines.}
We conduct comparisons with vanilla 3DGS~\cite{3dgs}, cascaded baselines, anti-aliasing techniques (\ie, Mip-Splatting~\cite{mipgs} and Analytic-Splatting~\cite{liang2025analytic}), and 3D super-resolution GaussianSR~\cite{yu2024gaussiansr}. 
Vanilla 3DGS and anti-aliasing baselines are trained on LR views and conduct HRNVS of target scale factors. For cascaded solutions, LR novel views are first rendered with vanilla 3DGS and then upsampled with upsamplers, including  Bicubic interpolation, arbitrary-scale ArbiSR~\cite{RealArbiSR}, and generative StableSR~\cite{stablesr}.  For simplicity, cascaded methods are denoted by corresponding upsamplers (\eg, Bicubic). We convert GaussianSR into an arbitrary-scale solution by training with random upscale factors.

\vspace{-4pt}\subsection{Quantitative Results}
Quantitative results are reported in Tab.~\ref{tab:quantitative}. Our method shows an overall superiority over baselines across four benchmarks, demonstrating effectiveness in rendering realistic HR outputs.  Specifically, vanilla 3DGS suffers from aliasing artifacts and obtains the worst performance. While Bicubic achieves favorable PSNR/SSIM, it lacks details and shows poor performance on FID.  ArbiSR and StableSR provide limited improvement on PSNR/SSIM due to altered contents (see Fig.~\ref {fig:mip360}).  Our method delivers superior perceptual quality over Mip-Splatting (by FID), and surpasses Analytic-Splatting in reconstructing high-fidelity results (\eg, a PSNR gain of 1.81 dB on Mip-NeRF360 at $\times 8$). Compared to GaussianSR, our method is consistently superior over all metrics.

\vspace{-4pt}\subsection{Qualitative Results}\label{sec:qualitative}
We provide qualitative comparisons on Mip-NeRF360 in Fig.~\ref{fig:mip360}, where full-resolution images are included as reference.  As can be seen, 3DGS suffers from aliasing artifacts, leading to severely degraded structures and erosion effects (\eg, the grass in the 1st row). Bicubic and Mip-Splatting produce over-smoothed outputs (\eg, the bonsai in the 3rd row). ArbiSR introduces distorted textures that compromise the visual quality. Despite StableSR providing visually appealing results, it alters the content and deviates from the original LR input views. Analytic-Splatting and GaussianSR hallucinate high-frequency artifacts that resemble textures.  In contrast, our method consistently delivers high-quality results, while faithfully preserving the original structures. We include more qualitative comparisons in the supplementary materials.

\begin{table}[t]
\centering
\Large
\resizebox{1\linewidth}{!}{
\begin{tabular}{l|ccc|cc}
\toprule
 Method &  \makecell{Rendering \\(ms)} &  \makecell{Throughput\\(FPS)} &  \makecell{Storage Size\\(GB)} & \makecell{Training \\(min)} &  \makecell{Memory \\(MB)}  \\
\midrule
3DGS                & 13    & 74   & 0.99  &  10  & 718  \\
Bicubic             & 34    & 29   & -   & - & - \\
ArbiSR              & 3225  & 0.31 & -    & - & -  \\
StableSR            & 10890 & 0.13 & -  & - & -  \\
Mip       & 19    & 52   & 0.99 &  12  & 858  \\
Analytic  & 33    & 30   & 1.20 & 29  & 822   \\
GaussianSR          & 8     & 126  & 0.56  &  256 & 8274 \\
Ours      & 12    & 85   & 0.79  &  57  & 7160 \\
\bottomrule
\end{tabular}
}
\caption{Efficiency analysis measured on a single NVIDIA A6000 GPU. Rendering time and throughput are evaluated at an output resolution of $1920 \times 1080$.  Mip and Analytic denote Mip-Splatting and Analytic-Splatting, respectively.}\vspace{-0.1in}
\label{tab:efficiency}
\end{table}  

\begin{table*}[t]
\centering
\Large
\resizebox{1\linewidth}{!}{%
\begin{tabular}{c|ccc|ccc|ccc|ccc|ccc}
\toprule
 \multirow{2}{*}{Mip-NeRF360} & \multicolumn{3}{c|}{$\times 2$} & \multicolumn{3}{c|}{$\times 4$}  &  \multicolumn{3}{c|}{$\times 8$} &   \multicolumn{3}{c|}{$\times 3.5$} &   \multicolumn{3}{c}{$\times 5.7$} \\
\cmidrule{2-16} 
&   PSNR$\uparrow$  &   SSIM$\uparrow$  &  FID$\downarrow$   
   &  PSNR$\uparrow$  &   SSIM$\uparrow$  &  FID$\downarrow$   
   & PSNR$\uparrow$  &   SSIM$\uparrow$  &  FID$\downarrow$   
    & PSNR$\uparrow$  &   SSIM$\uparrow$  &  FID$\downarrow$   
    & PSNR$\uparrow$  &   SSIM$\uparrow$  &  FID$\downarrow$   \\
\midrule
w/o 3D-SASF   & \cellcolor{yellow!30}26.13	  & 0.757	 &	41.576  &	24.85	  & 0.685  &	43.804  &	24.39  &	0.697    &	43.888	  &25.04	  & 0.693	  &	43.149  &	24.57  &	0.681   & 	43.881  \\
w/o 2D-SAMF &  25.53 & 0.736 & \cellcolor{yellow!30}36.858  & 24.83 & 0.679 &39.613& 24.61 & 0.703 & 39.774 & 24.93 & 0.684 & 39.006 & 24.71 & 0.682 & 39.610   \\					
w/o scale-aware  & 26.10 &0.758 &37.052 & \cellcolor{yellow!30}25.17 & 0.698 & \cellcolor{yellow!30}39.525  & \cellcolor{red!30}24.87 & 0.721 & \cellcolor{yellow!30}39.635	 & \cellcolor{yellow!30}25.30 & \cellcolor{red!30}0.703 & \cellcolor{yellow!30}38.901 & \cellcolor{red!30}25.00 & \cellcolor{yellow!30}0.700 & \cellcolor{yellow!30}39.523  \\	
     \midrule
 w/o PSR &  26.03	 &0.753 &	37.921	 & 24.51	 &0.663 &39.748	 & 23.91	 &0.664	 &39.873	 & 24.74	 &0.674	 &39.124	 &  24.14	 &0.651 &	39.797  \\
      \midrule 
w/o GPO & 25.23	  & 0.667   & 99.685	  &24.51	  &0.656  &	101.618	  &24.27  &	0.714	  & 	101.451	 	  &24.61	  & 0.652	  &	101.204	  & 24.37	  & 0.678	  & 101.530 \\	
w/o Orth &25.97&	\cellcolor{yellow!30}0.764	&39.392	&25.11	& \cellcolor{red!30}0.713	&	40.688	&24.85	&\cellcolor{red!30}0.741	&	40.573	&25.24	&0.717	&	40.106	& 24.96	&0.718	&	40.633 \\	
Pseudo HR  & 23.96	 &0.593	 &	111.149	 &23.36	 &0.600	 &111.574	 &23.19	 &0.683	 &	111.215 &	23.44 &0.592	 &	111.584	 &23.26	 &0.636	 &	111.526 \\
SDS loss &23.52	&0.698	&	72.640&	22.91	&0.655	&	75.134	&22.71&	0.697	&75.830	&22.99	&0.657	&	74.753	&22.79	&0.667	&	75.336 \\
\midrule
Ours & \cellcolor{red!30}26.23	& \cellcolor{red!30}0.764	& 	\cellcolor{red!30}36.524	& \cellcolor{red!30}25.18& \cellcolor{yellow!30}0.703& \cellcolor{red!30}38.526	& 	\cellcolor{yellow!30}24.85& 	\cellcolor{yellow!30}0.725&  \cellcolor{red!30}38.645 & \cellcolor{red!30}25.32	&  \cellcolor{red!30}0.709&  \cellcolor{red!30}37.924		& \cellcolor{yellow!30}24.99	& \cellcolor{red!30}0.704	& \cellcolor{red!30}38.606 \\
\bottomrule
\end{tabular}  
}\vspace{-4pt}
\caption{Ablation studies on Mip-NeRF360 dataset, where the best and second best results are highlighted with \colorbox{red!30}{red} and \colorbox{yellow!30}{yellow}, respectively.  3D-SASF, 2D-SAMF, PSR, GPO, Orth denote the 3D scale-aware smoothing filter, 2D scale-aware Mip filter, progressive super-resolving, generative prior-guided optimization and orthogonal reference refinement, respectively.} 
\label{tab:ab_mip}
\end{table*}

\subsection{Efficiency Analysis}\label{sec:efficiency}
Efficiency comparisons on rendering time, throughput, storage cost, training time, and  GPU memory usage are included in Tab.~\ref{tab:efficiency}. Our method introduces no extra computational overhead during rendering, achieving a throughput of 85 FPS for 1080p rendering, significantly faster than cascaded solutions (\eg, 908$\times$ faster than StableSR).  With fewer 3D Gaussians (as reflected by reduced storage size), GaussianSR and our method naturally render faster over Mip-Splatting and Analytic-Splatting. Notably, our method still delivers superior performance in Tab.~\ref{tab:quantitative} (\eg, a PSNR gain of 2.13 dB over Mip-Splatting on Blender at $\times 3.5$), highlighting its effectiveness in learning compact and expressive 3D scene representations. During training, generative prior-guided optimization inevitably increases the training time and GPU memory usage. However, the overhead remains affordable on commonly used GPUs. For instance, training takes only 57 min per scene with approximately 7 GB of memory usage.

\begin{figure}[t]
     \centering
        \includegraphics[width=1\linewidth,clip=true, trim=0 16pt 12pt 0]{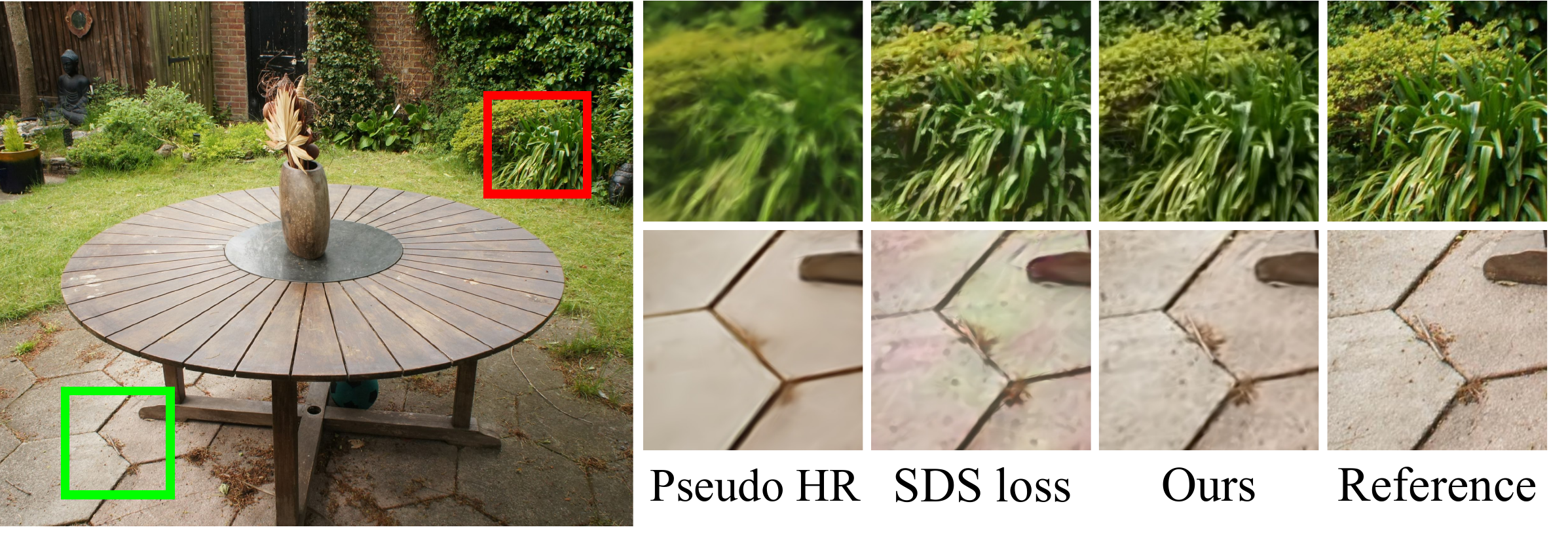}  \vspace{-0.18in}
        \caption{Qualitative comparisons with typical optimization strategies at the scale factor of $\times 4$. Replacing our generative prior-guided optimization with pseudo HR supervision and SDS loss results in blurriness and color distortion.}   \vspace{-0.1in}
      \label{fig:ablation_dis}
\end{figure}

\vspace{-4pt}\subsection{Ablation Studies}\label{sec:ablation} 
We assess each component's contribution by removing it. Then we examine the effectiveness of scale-awareness in the 2D scale-aware Mip filter. We further compare our generative prior-guided optimization with typical optimization strategies. Ablation studies are conducted on both Mip-NeRF360 and Tanks\&Temples (see supplementary materials) for generality.

\noindent \textbf{3D Scale-Aware Smooth Filter.} As shown in Tab.~\ref{tab:ab_mip}, removing 3D scale-aware smooth filter (denoted as w/o 3D-SASF) results in degraded performance. For instance, it leads to a PSNR drop of 0.46 dB on Mip-NeRF360 at $\times 8$.

\noindent \textbf{2D Scale-Aware Mip Filter.} As shown in Tab.~\ref{tab:ab_mip}, removing 2D scale-aware Mip filter (denoted as w/o 2D-SAMF) leads to a notable performance drop. For instance, the PSNR shows a drop of 0.7 dB at $\times 2$.   Unlike the vanilla 2D Mip filter~\cite{mipgs}, our proposed filter adaptively adjusts the integration window based on the scale factor, enabling more accurate pixel shading (as outlined Sec.\ref{sec:sar}). To investigate the effectiveness of scale-awareness, we replace 2D scale-aware Mip filter with the fixed-window 2D Mip filter (denoted as w/o SA).   As shown in Tab.~\ref{tab:ab_mip}, this results in suboptimal performance, especially in realism-based FID.  Visual comparisons are included in the supplementary materials.

\noindent \textbf{Progressive Super-Resolving.} To demonstrate the effectiveness of progressive super-resolving, we replace it with a mix-training strategy of random scale factors (denoted as  w/o PSR). As shown in the 4th row of Tab.~\ref{tab:ab_mip}, training without progressive super-resolving leads to suboptimal results.

\noindent \textbf{Generative Prior-Guided Optimization.} As shown in Tab.~\ref{tab:ab_mip}, removing generative prior-guided optimization (w/o GPO) leads to a significant performance drop (\eg, a PSNR drop of 1 dB at $\times 2$). Without generative priors, outputs exhibit reduced realism, as evidenced by increased FID scores.

\noindent \textbf{Comparison of Typical Optimization Strategies.} We further compare our GPO with two widely used optimization strategies: \textit{(1) pseudo HR supervision}, which imposes pixel-wise loss with pseudo HR views generated by pretrained SR models~\cite{xie2024supergs,srgs}; and \textit{(2) Score Distillation Sampling (SDS) loss}, which leverages generative priors via a denoising process~\cite{pooledreamfusion}. As shown in Tab.~\ref{tab:ab_mip}, pseudo HR supervision yields suboptimal results, particularly in perceptual realism (as reflected by FID), due to inconsistencies across views that hinder detail reconstruction and cause blur (see Fig.~\ref{fig:ablation_dis}). SDS loss also shows notable drops in PSNR/SSIM, exhibiting color distortions in Fig.~\ref{fig:ablation_dis}. In contrast, our GPO consistently delivers superior performance, effectively refining details and keeping fidelity.

\vspace{-4pt}\section{Conclusion}\label{sec:conclusion} 
This paper introduces an integrated framework for arbitrary-scale 3D Gaussian super-resolution. To address the key challenges, our framework incorporates three core components: scale-aware rendering, which enables adaptive HR rendering across varying scale factors; generative prior-guided optimization, which constrains fine details in the absence of ground-truth HR views; and progressive super-resolving, which gradually super-resolves 3D models to preserve content consistency across varying scale factors. Experimental results on four benchmarks demonstrate the effectiveness of our method in rendering high-quality arbitrary-scale HR views with a single 3D model. Ablation studies further verify the effectiveness of each component in our framework.

\bibliography{aaai2026}

\clearpage
\appendix
\onecolumn 
\input{supp}

\end{document}

%% file: supp.tex
\begin{algorithm*}[!t] 
\caption{Training Procedure of Arbitrary-Scale 3D Gaussian Super-Resolution}\label{code:training}
\textbf{Input}:  3D Gaussian primitives $\{\boldsymbol{G}^{3D}_i\}_{i=1}^N$, opacity $\{\boldsymbol{\alpha}_i\}_{i=1}^N$, color $\{\boldsymbol{c}_i\}_{i=1}^N$, input LR views $\{I_{LR}\}_{k=1}^K$, orthogonal view indicator $\{\mathbb{I}_{ortho}\}_{k=1}^K$, focal length $\{f_k\}_{k=1}^K$, camera depth $\{d_k\}_{k=1}^K$, target scale factors $\{s_k\}_{k=1}^K$  \\
\textbf{Output}: Optimized 3D Gaussian primitives $\{\boldsymbol{G}^{3D}_i\}_{i=1}^N$, opacity $\{\boldsymbol{\alpha}_i\}_{i=1}^N$, color $\{\boldsymbol{c}_i\}_{i=1}^N$ \\
\begin{algorithmic}[1] 
\STATE Initialize $\{\boldsymbol{G}^{3D}_i\}_{i=1}^N$;
\WHILE{trainingstage $t$ $\leq$ MaxStage $T$} 
\WHILE{iteration $\leq$ MaxIteration}
\STATE $I_{LR}$, $\mathbb{I}_{ortho}$, $f_k$, $d_k$, $s_k$ $\gets$ SampleTrainingView();
\STATE \hfill \textit{/* Sec.4.1: Scale-Aware Rendering */}
\STATE \hfill \textit{/* 3D Scale-Aware Smoothing Filter */}
\STATE $\{\boldsymbol{G}^{3D}_i\}_{i=1}^N \gets$  3DScaleAwareSmoothingFilter($\{\boldsymbol{G}^{3D}_i\}_{i=1}^N$, $\mathbb{I}_{ortho}$, $\{f_k\}_{k=1}^K$,  $\{d_k\}_{k=1}^K$, $\{s_k\}_{k=1}^K$);

   \STATE $\{\boldsymbol{G}^{2D}_i\}_{i=1}^N \gets$ Projection($\{\boldsymbol{G}^{3D}_i\}_{i=1}^N, \boldsymbol{P},\boldsymbol{W},\boldsymbol{J}$);\hfill \textit{/* 
Projecting 3D to 2D */}  
\STATE $\{\boldsymbol{G}^{2D}_i\}_{i=1}^N \gets$ 2DScaleAwareMipFilter($\{\boldsymbol{G}^{2D}_i\}_{i=1}^N$, $s_k$);\hfill \textit{/* 2D Scale-Aware Mip Filter */}

\STATE $I^t_{SR} \gets$ Pixelshading($\{\boldsymbol{G}^{2D}_i\}_{i=1}^N,  \{\boldsymbol{\alpha}_i\}_{i=1}^N, \{\boldsymbol{c}_i\}_{i=1}^N$);
\STATE    \hfill \textit{/* Sec.4.2: Generative Prior-Guided Optimization */}
\STATE $z_{SR}^{\hat{n}}$ $\gets$ DiffusionDenoise($I_{S R}^t, z, \hat{n}$);
\STATE $z_{LR}^n$ $\gets$ DiffusionDenoise($I_{LR}^t, z, n$);
\STATE $\mathcal{L}_{\mathrm{LDS}}$ $\gets$ LatentDistillationSamplingLoss($z_{SR}^{\hat{n}}$, $z_{LR}^n$);\hfill \textit{/* Generative Latent Distillation */}
\STATE \hfill \textit{/* Orthogonal Reference Refinement */}
\IF{$\mathbb{I}_{ortho}$} 
    \STATE  $I^t_{Ref}$ $\gets$ DiffusionDecoding($z^n_{LR},n$);
    \STATE $\mathcal{L}_{tex}$ $\gets$ TextureLoss($I^t_{SR}$, $I^t_{Ref}$);
\ELSE
    \STATE $\mathcal{L}_{tex}=0$;
\ENDIF
\STATE \hfill \textit{/* Sec.4.3: Progressive Super-Resolving*/}
\STATE $I^{t-1}_{SR}$ $\gets$ Rasterize($\{\boldsymbol{G}^{3D}_i\}_{i=1}^N$, $\{\boldsymbol{\alpha}_i\}_{i=1}^N$, $\{\boldsymbol{c}_i\}_{i=1}^N$, $f_k$, $d_k$, $s^{t-1}_k$); \hfill \textit{/* Lower-resolution View From Last Stage*/}
\STATE $\mathcal{L}_{str}$ $\gets$ StructureLoss($I^{t-1}_{SR}$, $I^t_{SR}$);
\STATE $\mathcal{L}$ $\gets$ WeightedLossSum($\mathcal{L}_{LDS}$, $\mathcal{L}_{tex}$, $\mathcal{L}_{str}$)
\STATE  $\{\boldsymbol{G}^{3D}_i\}_{i=1}^N$,  $\{\boldsymbol{\alpha}_i\}_{i=1}^N$, $\{\boldsymbol{c}_i\}_{i=1}^N$ $\gets$ AdamOptimize($\nabla\mathcal{L}$)
\ENDWHILE
\ENDWHILE
\end{algorithmic}
\end{algorithm*}

\section{Training Procedure Algorithm}\label{sec:code}
The pseudo-code of the training procedure is provided in Algorithm~\ref{code:training}. Our framework consists of three components: scale-aware rendering, generative prior-guided optimization, and progressive super-resolving. During training, the progressive super-resolving divides the whole training process into multiple stages. For each stage, scale-aware rendering is applied to synthesize aliasing-free novel views, while generative prior-guided optimization leverages generative priors to constrain fine details in the novel views. During inference, only scale-aware rendering is utilized.

\section{Latent Distillation Loss}\label{sec:lds}
The proposed Latent Distillation Sampling (LDS) loss can be regarded as a task‑specific variant of Score Distillation Sampling (SDS)~\cite{pooledreamfusion}, which is tailored for the super-resolution scenario. In this section, we show the derivation of LDS loss from the standard KL term. 

We begin with a standard KL term in SDS loss, which aligns the generated output with the distribution dictated by an external condition (\eg, a text prompt):
 \begin{equation}\label{eq: classical kl}
\operatorname{KL}\left(q\left(z_n \mid x=g(\theta)\right) \| p_\phi\left(z_n \mid y\right)\right) 
= \mathbb{E}_\epsilon\left[\log q\left(z_n \mid x=g(\theta)\right)-\log p_\phi\left(z_n \mid y\right)\right],
\end{equation}
where $z_n$, $x$ and $y$ denote the noisy latent at timestep $n$, output generated by $g(\theta)$, and external conditioning input, respectively. Equation~\ref{eq: classical kl} aims to optimize $x$ so that after Gaussian corruptions, its latent $z_n$ falls into the externally condition distribution. Given a super-resolution task where low-resolution (LR) views are available, we aim to learn a super-resolved view so that it remains consistent with LR input while yielding sharper details. A natural way is to replace the externally conditioned distribution with LR‑conditioned high-quality distribution, leading to the KL term below:
 \begin{equation}\label{eq: revised kl}
\operatorname{KL}\left(p_\phi\left(z_{\hat{n}} \mid I_{SR}=g(\theta)\right) \| p_\phi\left(z_n \mid I_{LR} \right)\right) \\
 =\mathbb{E}_{\hat{n}}\left[\log p_\phi\left(z_{\hat{n}} \mid I_{SR}\right)-\log p_\phi\left(z_n \mid I_{LR}\right)\right],
 \end{equation}
where $g(\theta)$ denote rendering super-resolved $I_{SR}$ (we omit $t$ in Eq.~\ref{eq:loss_tex}\footnote{To differentiate from this supplementary material, we use abbreviations to denote sections, tables, and figures in the main texts (\ie., ``Sec.'' for sections, ``Tab.'' for tables, and ``Fig.'' for figures)} for simplicity) with learnable parameters $\theta$ of 3D Gaussian primitives. We set $\hat{n}>n$ to ensure the high-quality distribution conditioned on LR, and encourage SR output to further move towards this distribution. The gradient of the KL term in Equation~\ref{eq: revised kl} is:
{\small
\begin{equation}
 \nabla_\theta \operatorname{KL}\left(p_\phi\left(z_{\hat{n}} \mid I_{SR} =g(\theta)\right) \| p_\phi\left(z_n \mid I_{LR} \right)\right)   \\
=\mathbb{E}_{\hat{n}}[\underbrace{\nabla_\theta \log p_\phi\left(z_{\hat{n}} \mid I_{SR}\right)}_{\text {(A) }}-\underbrace{\nabla_\theta \log p_\phi\left(z_n \mid I_{LR}\right)}_{\text {(B) }}].
\end{equation}
}
Applying the chain rule to the term A and B yields a common factor of the form  $\nabla_z \log p_\phi\left(z \mid I \right)\frac{\partial z}{\partial \theta}$, where $\nabla_z \log p_\phi\left(z \mid I \right)$ relates to the unknown noise residual, and can be approximated with the noise $s_\phi\left(z \mid I\right)$ predicted by UNet. This provides a low-variance gradient estimation as follows:
\begin{equation}\label{eq:a}
(A) \approx  s_\phi\left(z_{\hat{n}}\mid I_{SR}\right) \frac{\partial z_{\hat{n}}}{\partial \theta} 
=\alpha s_\phi\left(z_{\hat{n}} \mid I_{SR}\right) \frac{\partial I_{SR}}{\partial \theta} \\
=-\frac{\alpha_{\hat{n}}}{\sigma_{\hat{n}}} \epsilon_\phi\left(z_{\hat{n}} \mid I_{SR}\right) \frac{\partial I_{SR}} {\partial \theta},
\end{equation}
\begin{equation}\label{eq:b}
(B) \approx  s_\phi\left(z_n \mid I_{LR}\right) \frac{\partial z_n}{\partial \theta} 
=\alpha s_\phi\left(z_n \mid I_{LR}\right) \frac{\partial I_{LR}}{\partial \theta}\\
=-\frac{\alpha_n}{\sigma_n} \epsilon_\phi\left(z_n \mid I_{LR}\right) \frac{\partial I_{LR}} {\partial \theta}.
\end{equation}
In practice, $n$ is a constant, $I_{LR}$ is fixed and independent of $\theta$. Consequently, only $I_{SR}$ contributes to the gradient of the LDS loss, which is derived as follows:
\begin{equation}
\begin{aligned}
\nabla_\theta \mathcal{L}_{\mathrm{LDS}} & =\mathbb{E}_{\hat{n}, z_{\hat{n}} \mid I_{SR}}\left[\frac{\sigma_{\hat{n}}}{\alpha_{\hat{n}}} \nabla_\theta \operatorname{KL}\left(p_\phi\left(z_{\hat{n}} \mid I_{SR}=g(\theta)\right) \| p_\phi\left(z_n \mid I_{LR} \right)\right)\right] \\
&=\mathbb{E}_{\hat{n}}\left[w(\hat{n}) \cdot  \left(-\epsilon_\phi\left(z_{\hat{n}} \mid I_{SR}\right) \frac{\partial I_{SR}} {\partial \theta}  + \epsilon_\phi\left(z_n \mid I_{LR}\right) \frac{\partial I_{LR}} {\partial \theta} \right) \right] \\
&= \mathbb{E}_{\hat{n}}\left[w(\hat{n}) \cdot \left (\epsilon_\phi\left(z_{\hat{n}} ; I^t_{SR}, \hat{n}\right)-\epsilon_\phi\left(z_n;  I_{LR}, n\right)\right) \frac{\partial I_{SR}}{\partial \theta}\right], 
\end{aligned}
\end{equation}
where $w(\hat{n})$ is a weighting term depending on the timestep $\hat{n}$. $\epsilon_\phi(\cdot)$ indicates predicting the noise with UNet.

\begin{table*}[t]
\centering
\resizebox{0.7\linewidth}{!}{
\begin{tabular}{l|cc|cc|cc|cc}
\toprule
\multirow{2}{*}{Method} & \multicolumn{2}{c|}{Blender}  & \multicolumn{2}{c|}{Mip-NeRF360} & \multicolumn{2}{c|}{Tanks\&Temples} & \multicolumn{2}{c}{Deep Blending} \\
\cmidrule{2-9}
 & $\times 12$& $\times 16$ & $\times 12$& $\times 16$ &  $\times 12$& $\times 16$ & $\times 12$& $\times 16$  \\
\midrule
3DGS & 256.287 & 262.810  & 92.282 & 95.412  & 189.776 & 196.949  & 266.423 & 271.213\\
Bicubic  & 182.535 & 182.538  & 47.473 & 47.483  & 142.016 & 142.023 &  185.528 & 185.636\\
ArbiSR  &  171.130 & 169.713  & 54.522 & 54.416  & 155.999 & 152.882  & 199.136 & 198.439\\
StableSR  & \cellcolor{yellow!30}94.952 & \cellcolor{yellow!30}94.936 &  106.997 & 106.996  & \cellcolor{red!30}86.111 & \cellcolor{red!30}86.073  & \cellcolor{red!30}128.757 & \cellcolor{red!30}128.773\\
Mip-Splatting  & 113.081 & 113.060 &53.736 & 53.732 &99.837 & 99.923 &  189.625 & 189.735\\
Analytic-Splatting  & 147.154 & 147.488  & \cellcolor{red!30}35.363 & \cellcolor{red!30}32.368 & 145.031 & 145.490  & 218.129 & 218.452\\
GaussianSR  & 113.725 & 112.556  & 110.681 & 107.030 &  114.808 & 114.872  & 172.658 & 172.634\\
Ours   & \cellcolor{red!30}87.842 & \cellcolor{red!30}87.797   & \cellcolor{yellow!30}38.828 & \cellcolor{yellow!30}38.824  & \cellcolor{yellow!30}90.174 & \cellcolor{yellow!30}90.155 &  \cellcolor{yellow!30}154.890 & \cellcolor{yellow!30}154.887\\
\bottomrule
\end{tabular}
}\vspace{-2pt}
\caption{Quantitative comparison on higher scale factors, where the best and second-best results are highlighted in \colorbox{red!30}{red} and \colorbox{yellow!30}{yellow}, respectively. }
\label{tab:qualiclip}
\end{table*}

\begin{table*}[t]
\Large
\resizebox{1\linewidth}{!}{
\begin{tabular}{l|cccc|cccc|cccc|cccc}
\toprule
\textbf{Method} & \multicolumn{4}{c|}{Blender} & \multicolumn{4}{c}{Mip-NeRF360} & \multicolumn{4}{c|}{Tanks\&Temples} & \multicolumn{4}{c}{Deep Blending}\\  
            & PSNR↑ & SSIM↑ & FID↓ & LPIPS↓ & PSNR↑ & SSIM↑ & FID↓ & LPIPS↓ & PSNR↑ & SSIM↑ & FID↓ & LPIPS↓ & PSNR↑ & SSIM↑ & FID↓ & LPIPS↓\\
\midrule
3DGS            & 21.56 & 0.878 & 52.604 & 0.070 & \cellcolor{yellow!30}27.79 & \cellcolor{red!30}0.846& 33.176 & 0.103 & 24.44	&0.893	&49.569 &0.042	 & 27.87	& 0.874	&104.162 & \cellcolor{yellow!30}0.065	\\
Mip-Splatting   & 24.88 & 0.912 & \cellcolor{yellow!30}38.399 & 0.046 &  \cellcolor{red!30}27.80 & 0.837 & \cellcolor{red!30}28.037 & \cellcolor{yellow!30}0.090  & \cellcolor{red!30}24.69 & 	\cellcolor{yellow!30}0.903	 & \cellcolor{yellow!30}48.273 & 0.039	  & 28.05	 & 0.879	 & 	\cellcolor{yellow!30}100.688 & 0.070 \\
Analytic-Splatting & \cellcolor{yellow!30}27.16&	\cellcolor{yellow!30}0.941	&40.753&\cellcolor{yellow!30}0.029	 & 27.41	 & 0.826& 	35.888  & 	0.100 &\cellcolor{yellow!30}24.56	 & 0.895	 & 51.855  & 	\cellcolor{yellow!30}0.037& 27.83	 & 0.873	 & 	111.294 & 0.070 \\
GaussianSR    & 22.96	&0.877&	62.846 & 	0.060 & 26.10	 &0.734	 &96.792   &0.266	& 24.34 & 	0.872& 59.343  & 	0.048	& \cellcolor{yellow!30}28.14	& \cellcolor{yellow!30}0.886	& 104.917& 0.082	\\
\textbf{Ours}  & \cellcolor{red!30}27.42	& \cellcolor{red!30}0.945 & \cellcolor{red!30}29.185 & \cellcolor{red!30}0.029  & 27.52	 & \cellcolor{yellow!30}0.844  &	\cellcolor{yellow!30}29.547  &	\cellcolor{red!30}0.089 & 24.48	 & \cellcolor{red!30}0.909	& \cellcolor{red!30}47.020  & \cellcolor{red!30}0.036	 &  \cellcolor{red!30}28.45	 & \cellcolor{red!30}0.896  & \cellcolor{red!30}83.112  & 	\cellcolor{red!30}0.060	 \\
\bottomrule
\end{tabular}
}\vspace{-2pt}
\caption{Quantitative comparisons on scale factor of $\times 1$,  where the best and second-best results are highlighted in \colorbox{red!30}{red} and \colorbox{yellow!30}{yellow}, respectively.}
\label{tab:quanlix1}
\end{table*}

  \begin{figure*}[t]
    \centering
    \includegraphics[width=0.95\linewidth,clip=true, trim=18pt 20pt 2pt  0 ]{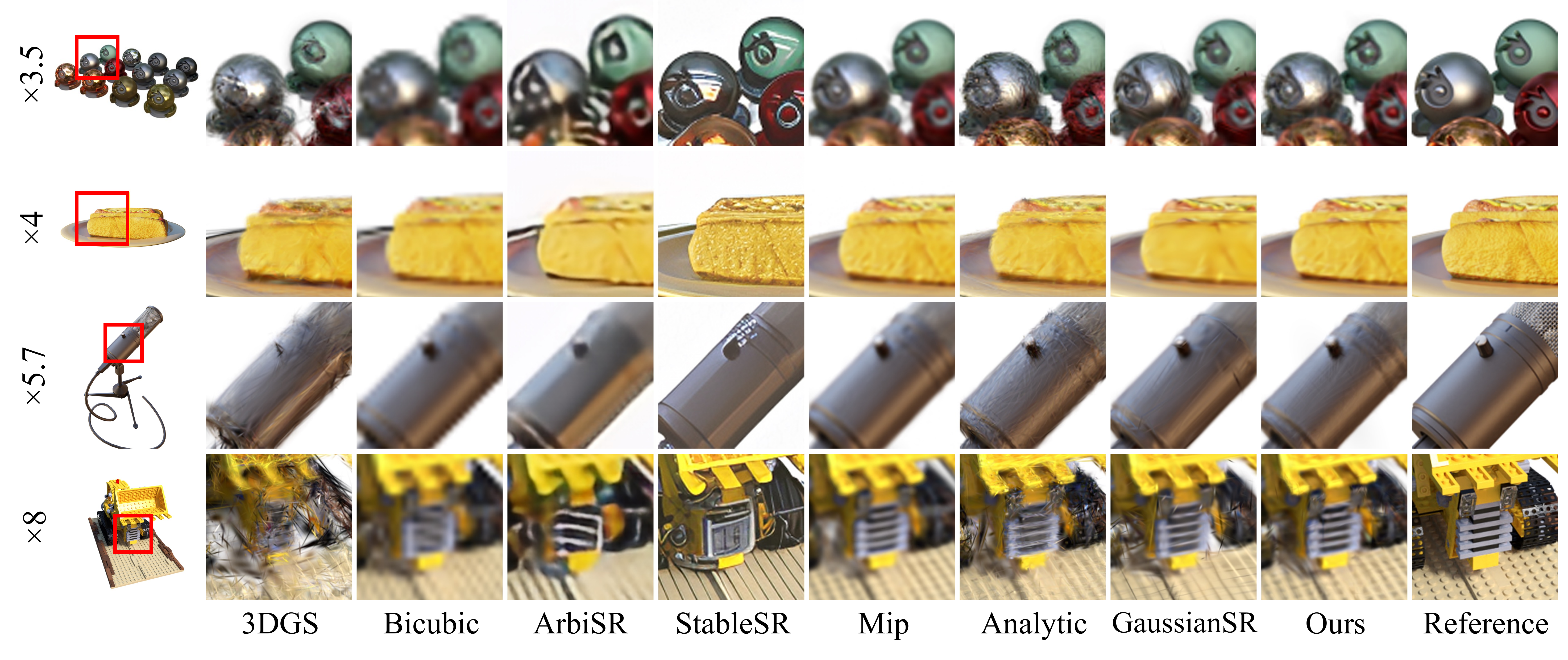} \vspace{-0.1in}
    \caption{Qualitative comparisons on Blender~\cite{nerf} dataset, where Mip and Analytic denote Mip-Splatting and Analytic-Splatting, respectively.  The proposed method effectively enhances fine details while avoiding visually unsatisfying artifacts (\eg, aliasing artifacts of 3DGS in the 1st column). Compared to cascaded solutions, our method preserves content integrity, achieving realistic and visually pleasant renderings. } 
    \label{fig:blender_apx}
\end{figure*}

\section{Generalization Ability Evaluation}\label{sec:generalization ability}
\subsection{Performance on Higher Scale Factors}\label{sec: x12}
We set the maximum scale factor $s_{max}=8$ during training. To evaluate the generalization ability beyond this range, we test with $\times 12$ and  $\times 16$. Since no pair-wise ground truth is available at these scales, we evaluate the performance with FID, using the original full-resolution dataset as the reference distribution.   As shown in Table~\ref{tab:qualiclip}, StableSR shows overall superior performance due to its highly realistic outputs, although it alters scene content (as discussed in Sec.~\ref{sec:qualitative}). Our method demonstrates consistently competitive performance, ranking among the top two methods. In addition, it significantly outperforms GaussianSR and anti-aliasing methods, exhibiting robust performance across higher scale factors.
 
\vspace{-4pt}\subsection{Performance on Novel View Synthesis ($s=1$)}\label{sec: x1}
Since we target 3DGS super-resolution, where the goal is to render higher resolution views with the scale factor $s>1$, we propose key components such as Generative Prior-guided Optimization and Scale-Aware Rendering for upsampling scenarios.  Given the task of novel view synthesis (NVS), instead of retraining the model with $s=1$, we consider it as an evaluation of the model's generalization ability at the unseen scale factor of $\times 1$, and directly render outputs with $s=1$. For the compared methods, please note that 3DGS, Mip-Splatting, and Analytic-Splatting are trained on with LR views (\ie, $\times 1$), whereas GaussianSR is trained under the arbitrary-scale configurations. As shown in Table~\ref{tab:quanlix1}, our method demonstrates strong generalization ability, surpassing all baselines on the Blender and Deep Blending dataset. For instance, our method obtains a PSNR of 0.26 dB over well-performing Analytic-Splatting on Blender. Additionally, the proposed method shows competitive performance with the NVS methods (\eg, Analytic-Splatting) on Mip-NeRF360 and Tanks\&Temples dataset. Compared with the arbitrary-scale GaussianSR, our method achieves consistently superior performance, especially on the synthetic Blender dataset, highlighting the effectiveness of our proposed components.

\begin{figure*}[t]
    \centering
    \includegraphics[width=0.95\linewidth,clip=true, trim=18pt 20pt 2pt  0]{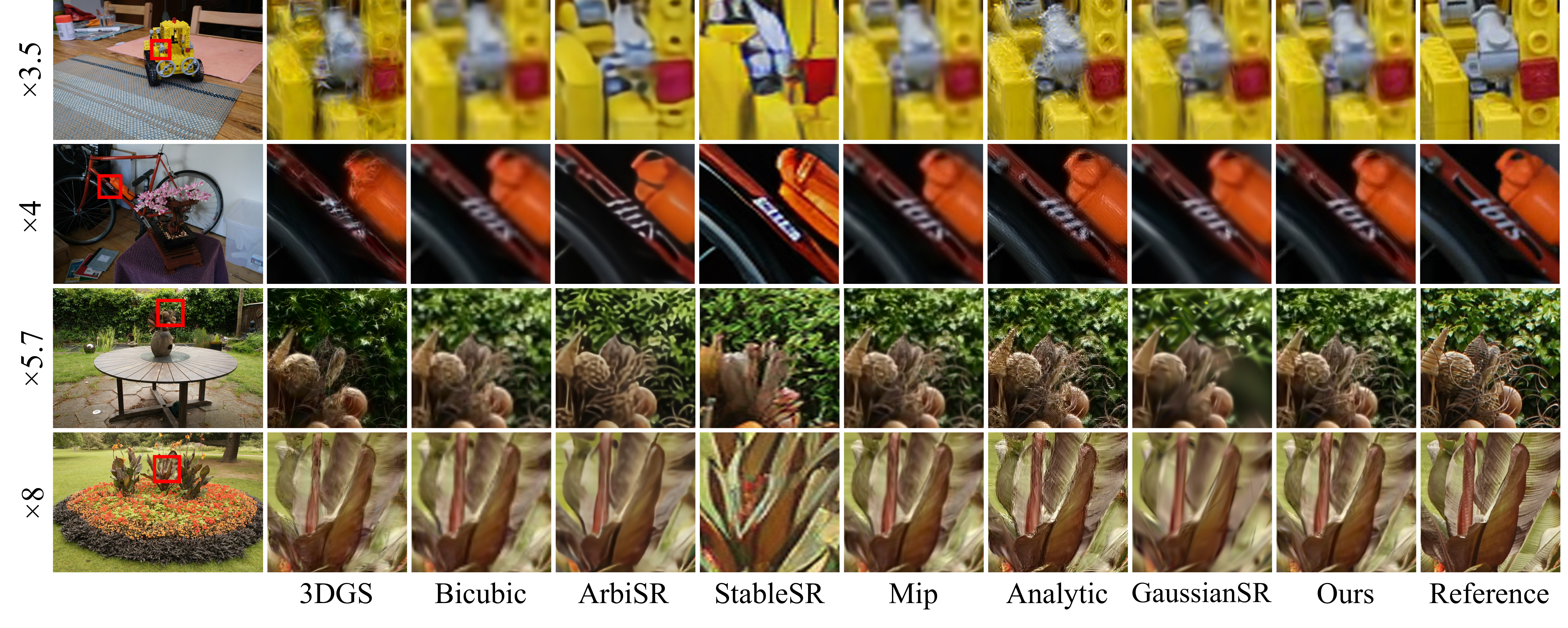} \vspace{-0.1in}
    \caption{Qualitative comparisons on the Mip-NeRF360~\cite{mipnerf360} dataset, where Mip and Analytic denote Mip-Splatting and Analytic-Splatting, respectively.  ArbiSR introduces distorted textures (\eg, the text shown in the 1st row), while StableSR changes the contents of the rendered view (\eg, changed appearance of lego in the 3rd row). Analytic-Splatting suffers from high-frequency artifacts. In contrast, the proposed method effectively renders high-fidelity results with rich details. } \vspace{-2pt}
    \label{fig:mip_apx}
\end{figure*}

\begin{table*}[t]
\centering
\Large
\resizebox{\textwidth}{!}{%
\begin{tabular}{l|c|c|ccc|ccc|ccc|ccc|ccc}
\toprule
\multirow{2}{*}{Stage} & \multirow{2}{*}{\makecell{Training Time \\(min)}} & \multirow{2}{*}{\makecell{GPU Memory \\(MB)}} & \multicolumn{3}{c|}{$\times 2$} & \multicolumn{3}{c|}{$\times 4$} & \multicolumn{3}{c|}{$\times 8$} & \multicolumn{3}{c|}{$\times 3.5$} & \multicolumn{3}{c}{$\times 5.7$} \\
 &  &  & PSNR$\uparrow$ & SSIM$\uparrow$ & FID$\downarrow$ & PSNR$\uparrow$ & SSIM$\uparrow$ & FID$\downarrow$ & PSNR$\uparrow$ & SSIM$\uparrow$ & FID$\downarrow$ &PSNR$\uparrow$ & SSIM$\uparrow$ & FID$\downarrow$  & PSNR$\uparrow$ & SSIM$\uparrow$ & FID$\downarrow$ \\
\midrule
Stage1 & 17 & 6897 & 25.31 & 0.920 & \textbf{65.896} & 23.93 & 0.894 & 90.908 & 23.03 & 0.875 & 94.445 & 24.14 & 0.899 & 89.703 & 23.48 & 0.883 & 93.287 \\
Stage2 & 18 & 6929 & 25.41 & 0.921 & 66.260 & 24.05 & 0.894 & 89.052 & 23.13 &  0.875 & 92.143 & 24.26 & 0.900 & 88.044 & 23.59 & 0.883 & 91.124 \\
 Stage3 & 22 & 7160 &  \textbf{25.60}	 & \textbf{0.925}	 & 	66.515	 & \textbf{24.32}	 & \textbf{0.899}	 & \textbf{86.270}	 & \textbf{23.40}	 & \textbf{0.879}	 & \textbf{87.984} & \textbf{24.53} & 	\textbf{0.904}	  & 	\textbf{85.788}	 & \textbf{23.87} & 	\textbf{0.888}	 &  \textbf{87.429} \\
\bottomrule
\end{tabular}%
}\vspace{-2pt}
\caption{Quantitative performance on Blender dataset across multiple training stages.}
\label{tab:ablation_stage}
\end{table*}

\begin{table*}[t]
\centering
\Large
\resizebox{1\linewidth}{!}{%
\begin{tabular}{c|ccc|ccc|ccc|ccc|ccc}
\toprule
 \multirow{2}{*}{Tanks\&Temples} & \multicolumn{3}{c|}{$\times 2$} & \multicolumn{3}{c|}{$\times 4$}  &  \multicolumn{3}{c|}{$\times 8$} &   \multicolumn{3}{c|}{$\times 3.5$} &   \multicolumn{3}{c}{$\times 5.7$} \\
\cmidrule{2-16} 
   &   PSNR$\uparrow$  &   SSIM$\uparrow$  &  FID$\downarrow$   
   &  PSNR$\uparrow$  &   SSIM$\uparrow$  &  FID$\downarrow$   
   & PSNR$\uparrow$  &   SSIM$\uparrow$  &  FID$\downarrow$   
    & PSNR$\uparrow$  &   SSIM$\uparrow$  &  FID$\downarrow$   
    & PSNR$\uparrow$  &   SSIM$\uparrow$  &  FID$\downarrow$   \\
     \midrule
w/o 3D-SASF  &  \cellcolor{red!30}23.11 & 	0.816	 &  \cellcolor{yellow!30}50.159	 & 20.87	 & 0.658 & 	86.167	 & 19.54	 & 0.566	 & 103.580	 & 21.26 & 0.687	 & 77.829	 & 20.16 & 	0.602	 & 97.379 \\
w/o 2D-SAMF &  21.14	 &0.738	 &	77.315	 & 19.97	 &0.608	 &	107.117	 & 19.12 &	0.539	 & 118.008	 & 20.19 &	0.631	 &	100.080	 & 19.56 &	0.565	 &	113.404  \\
w/o scale-aware  & 22.84	 &0.809	 &	50.754	 & 20.96	 & 0.661	 &	83.880	 &  19.84 &	0.578	 & 99.687	 & 21.28	 &0.686	 &	 76.918	 & 20.40 &	0.611	 &	 94.208  \\
\midrule
 w/o PSR & 22.68	 & 0.794	 &  60.371 &	20.05	 &0.597	 &	118.645	 &18.64 &	0.500	 &	139.571	  & 20.50	 &0.631	 &	106.419 & 	19.25 &	0.534	 &	132.967 \\
     \midrule  
w/o GPO &  22.81	& 0.781	&  62.446	&  21.05	&  0.652	&  90.483	&  19.94& 	0.581	& 104.621	& 21.36	& 0.674	& 84.315	& 20.50	& 0.609	& 100.216  \\
w/o Orth & \cellcolor{yellow!30}23.08 & \cellcolor{yellow!30}0.819 &  52.048 & \cellcolor{red!30}21.16 & \cellcolor{yellow!30}0.677 & 79.460 & \cellcolor{yellow!30}19.98 & \cellcolor{yellow!30}0.591 & 95.542  & \cellcolor{red!30}21.50 & \cellcolor{yellow!30}0.702 &  72.740 & \cellcolor{yellow!30}20.56 & \cellcolor{yellow!30}0.625 &  90.091 \\
Pseudo HR & 22.47	& 0.759	& 63.584	& 20.85	& 0.635	& 79.257	& 19.80	& 0.573 &	\cellcolor{yellow!30}89.672	& 21.13& 	0.656	& 76.234	& 20.33& 	0.596	& 	86.666\\
SDS loss & 21.89	&  0.794	& 57.122	& 20.29	&  0.659	& 	\cellcolor{yellow!30}76.844	&  19.28	&  0.582	&  \cellcolor{red!30}87.337	& 20.58	&  0.683	&  \cellcolor{yellow!30}72.491	& 19.78	& 0.612	& \cellcolor{red!30}83.896\\
\midrule
 Ours &  22.94	& \cellcolor{red!30}0.821	& \cellcolor{red!30}48.659	& \cellcolor{yellow!30}21.14	& \cellcolor{yellow!30}0.686	&	\cellcolor{red!30}75.591	& \cellcolor{red!30}19.99 &	\cellcolor{red!30}0.600	&	89.928	& \cellcolor{yellow!30}21.47	& \cellcolor{red!30}0.710	& 	\cellcolor{red!30}69.519	& \cellcolor{red!30}20.57	& \cellcolor{red!30}0.635	&	\cellcolor{yellow!30}85.057\\
\bottomrule
\end{tabular}  
} \vspace{-2pt}
\caption{Ablation studies on Tanks\&Temples~\cite{tandt} dataset, with the best and second best results highlighted in \colorbox{red!30}{red} and \colorbox{yellow!30}{yellow}, respectively.  GPO, PSR, 3D-SASF, and 2D-SAMF are short for generative prior-guided optimization, progressive super-resolving, 3D scale-aware smoothing filter, and 2D scale-aware Mip filter, respectively. } 
\label{tab:ab_tandt}
\end{table*}

\section{More Qualitative Comparisons}\label{sec: Qualitative}  
We present qualitative comparisons on Blender~\cite{nerf} and Mip-NeRF360~\cite{mipnerf360} dataset in Figure~\ref{fig:blender_apx} and Figure~\ref{fig:mip_apx}, respectively. Each result includes the corresponding full-resolution image from the dataset as a reference.
As can be seen, 3DGS suffers from aliasing artifacts, leading to eroded structures and high-frequency distortions, such as the lego in the 4th row of Figure~\ref{fig:blender_apx}. Bicubic interpolation, Mip-Splatting and GaussianSR reduce aliasing but produce over-smoothed outputs, lacking essential fine details for high-quality rendering (\eg, the microphone in the 3rd row of Figure~\ref{fig:blender_apx}). ArbiSR introduces noticeable texture distortions that degrade visual quality, as can be seen from the cannon part in the 1st row of Figure~\ref{fig:mip_apx}.  While StableSR generates visually appealing outputs, it synthesizes contents that deviate from the original LR input, as demonstrated by the texts in the 2nd row of Figure~\ref{fig:mip_apx}.  Analytic-Splatting suffers from ``thin Gaussians'', leading to the high-frequency artifacts that resemble textures (\eg, as shown in the 3rd row of Figure~\ref{fig:mip_apx}).  Among all the solutions, our method consistently produces high-quality results, while faithfully preserving structural integrity and original content, ensuring both realism and accuracy in the reconstructed views.

 \section{Stage-wise Performance Comparison}\label{sec: multistage}
To ensure consistency across different scale factors, we propose the progressive super-resolving strategy, which divides training into three stages (further detailed in Section~\ref{sec: Settings}). To demonstrate its effectiveness, we report the stage-wise performance on the Blender dataset in Table~\ref{tab:ablation_stage}.  As can be seen,  the progressive strategy leads to consistent performance improvements across stages, indicating its effectiveness in the task of arbitrary-scale 3D super-resolution.

\section{Ablation Studies}\label{sec: Ablation}
\subsection{Effectiveness of Each Component}\label{sec: comp}
We conduct ablation studies on both Mip-NeRF360 (see Sec.~\ref{sec:ablation}) and Tanks\&Temples for generality. For each component, we assess its contribution by removing it. We further compare our generative prior-guided optimization with typical optimization strategies to explore its potential. 

\noindent \textbf{3D Scale-Aware Smooth Filter.} As shown in Table~\ref{tab:ab_tandt}, discarding 3D scale-aware smooth filter (denoted as w/o 3D-SASF) leads to a significant performance drop, especially at the high upscale factors (\eg,  a PSNR drop of 0.45 dB at $\times 8$).

\noindent \textbf{2D Scale-Aware Mip Filter.} As shown in Table~\ref{tab:ab_tandt}, removing 2D scale-aware Mip filter (denoted as w/o 2D-SAMF) leads to suboptimal results (\eg,  a PSNR drop of 1.8 dB at $\times 2$). When replacing the 2D scale-aware Mip filter with a vanilla 2D Mip filter (denoted as w/o SA),  the model shows a notable performance drop, demonstrating the effectiveness by accumulating \begin{wrapfigure}{r}{0.47\linewidth}   
    \centering
 \includegraphics[width=0.9\linewidth,clip=true, trim=0 0.68in 12pt  0 ]{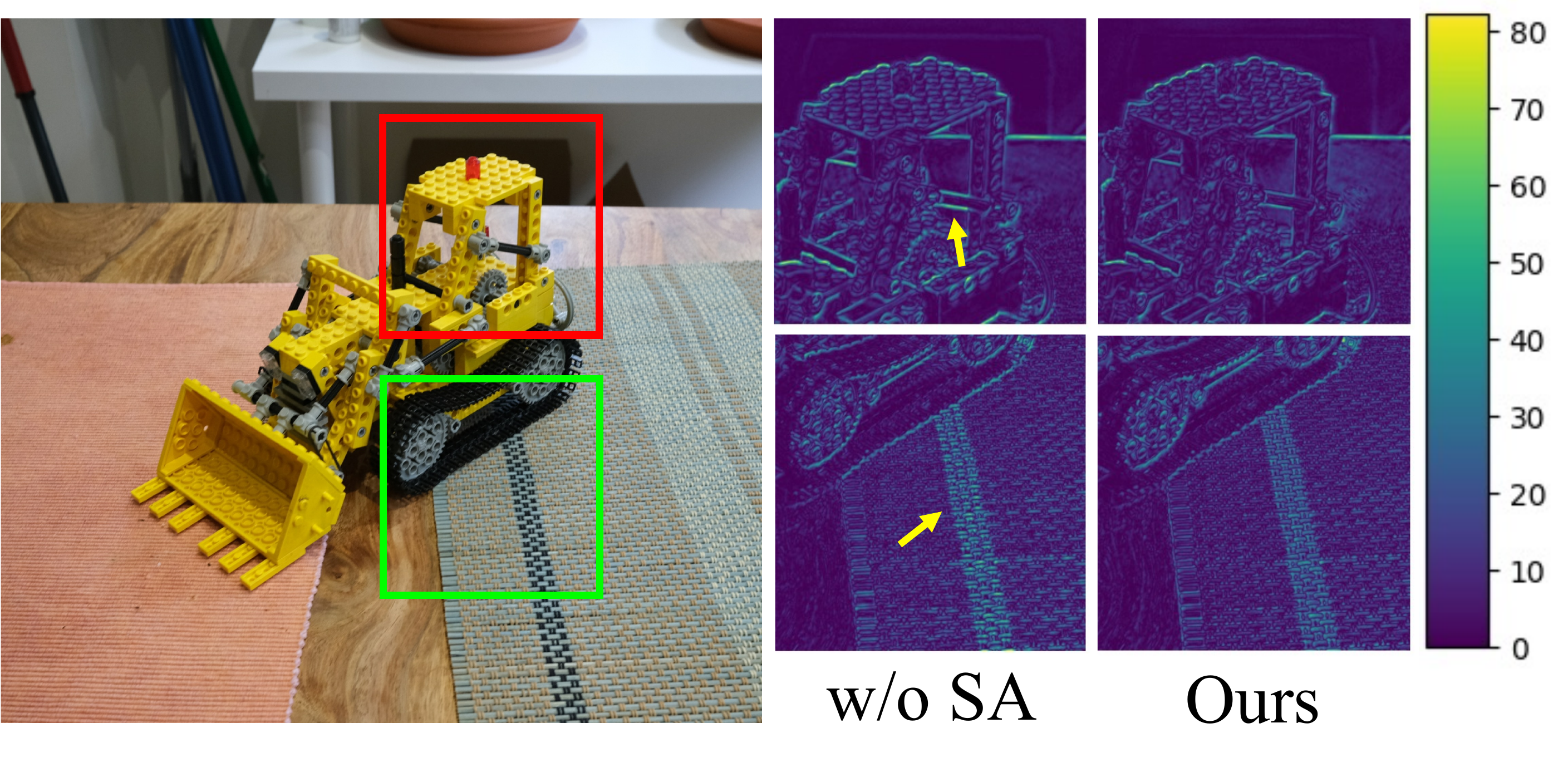}\vspace{-4pt}
    \caption{Qualitative comparisons on the effectiveness of the scale-aware (SA) design in 2D scale-aware Mip filter. We include the residual map between the rendered HR view and the reference view to highlight the differences.}
    \label{fig:ablation_sa}
\end{wrapfigure}  signals within an adaptive window size.   This is further evidenced by the residual comparisons in Figure~\ref{fig:ablation_sa}, where removing the scale-aware design results in structure distortions for high-frequency contents (e.g., edges indicated by the yellow arrows).
 
\noindent \textbf{Progressive Super-Resolving.} We replace the progressive super-resolving with a mix-training strategy of random scale factors (denoted as  w/o PSR). As shown in Table~\ref{tab:ab_tandt}, training without progressive super-resolving leads to suboptimal results. 

\noindent \textbf{Generative Prior-Guided Optimization.} As illustrated in Table~\ref{tab:ab_tandt}, removing generative prior-guided optimization (w/o GPO) leads to reduced realism (\eg, a PSNR drop of 1 dB at $\times 2$ and poor FID results). 

\noindent \textbf{Comparison of Typical Optimization Strategies.} We compare our GPO with representative optimization strategies: pseudo HR and Score Distillation Sampling loss. As shown in Table~\ref{tab:ab_tandt}, optimizing with pseudo HR results in suboptimal realism, which may come from the inconsistency across pseudo HR views. Meanwhile,  replacing GPO with SDS leads to a notable drop in PSNR/SSIM, indicating its ineffectiveness in preserving the fidelity.

\section{Experimental Settings}\label{sec: Settings}
\noindent \textbf{Datasets.}
Consistent with previous research~\cite{srgs,hu2024gaussiansr,yu2024gaussiansr}, we follow standard dataset splits for training and evaluation. For the Blender dataset~\cite{nerf}, we use 100 images for training and 200 images for testing. For each scene in the realistic datasets Mip-NeRF360~\cite{mipnerf360}, Tanks\&Temples~\cite{tandt}, and Deep Blending~\cite{db}, we adopt $\frac{7}{8}$ of the images for training and the remaining $\frac{1}{8}$ for testing. Unlike previous work~\cite{3dgs} that imposes size constraints on the longest side of images, we remove this constraint to enable high-resolution rendering of arbitrary scale factors. To generate low-resolution (LR) views a specified scale factor (\eg, $\times 3.5$), we adopt Bicubic interpolation to downsampling the original full-resolution images into LR based on the specified scale factor. Once obtaining the rendered high-resolution (HR) views, we evaluate the quality by comparing them against the full-resolution reference images.

\noindent \textbf{Implementation details.}  During training, we divide the entire process into 3 stages, corresponding to the maximum scale factors $s^t$ of 2, 4 and 8, respectively. Each stage takes 2,000 iterations for training to ensure converge and stable performance. We set the $\varepsilon$ to be 0.1. For the generative prior-guided optimization, we set the timestep $n=200$ to ensure high-quality latent variable, while timestep $\hat{n}$ is a randomly chosen between 0 and 200. The loss weight $\lambda_1$, $\lambda_2$ and $\lambda_3$ are set to 1 by default. The hyperparameter $\lambda$ in Eq.~\ref{eq:struct} is set to 0.5. All experiments are conducted with an A6000 GPU.